\documentclass[10pt,twocolumn,letterpaper]{article}

\usepackage[pagenumbers]{cvpr} 

\definecolor{cvprblue}{rgb}{0.21,0.49,0.74}
\usepackage[pagebackref,breaklinks,colorlinks,allcolors=cvprblue]{hyperref}

\usepackage{amsmath, amsfonts}  
\usepackage[accsupp]{axessibility}
\usepackage{booktabs}   
\usepackage{multirow}   
\usepackage{makecell}   
\usepackage{array}      
\usepackage{xcolor}     
\usepackage{colortbl}   
\usepackage{pifont} 
\usepackage{float}

\definecolor{lightmint}{RGB}{235,250,238}
\definecolor{lightblue}{RGB}{217,234,247}
\definecolor{lightcyan}{RGB}{221,243,250}
\definecolor{lightgreen}{RGB}{228,240,228}

\def\confName{CVPR}
\def\confYear{2026}

\title{Uncertainty-guided Compositional Alignment with Part-to-Whole Semantic Representativeness in Hyperbolic Vision-Language Models}

\author{
Hayeon Kim$^{1,*}$ \qquad
Ji Ha Jang$^{1,*}$ \qquad
Junghun James Kim$^{2}$ \qquad  
Se Young Chun$^{1,2,\dagger}$ \\
$^1$ Dept. of Electrical and Computer Engineering, $^2$ INMC \& IPAI \\
Seoul National University, Republic of Korea \\
{\tt\small \{khy5630, jeeit17, jonghean12, sychun\}@snu.ac.kr} \\
{\small $^*$Authors contributed equally. \quad $^\dagger$Corresponding author.}
}
\begin{document}
\maketitle


\begin{abstract}
While Vision-Language Models (VLMs) have achieved remarkable performance, their Euclidean embeddings remain limited in capturing hierarchical relationships such as part-to-whole or parent-child structures, and often face challenges in multi-object compositional scenarios. Hyperbolic VLMs mitigate this issue by better preserving hierarchical structures and modeling part-whole relations (\textit{i.e.}, whole scene and its part images) through entailment. However, existing approaches do not model that each part has a different level of semantic representativeness to the whole. We propose UNcertainty-guided Compositional Hyperbolic Alignment (UNCHA) for enhancing hyperbolic VLMs. UNCHA models part-to-whole semantic representativeness with hyperbolic uncertainty, by assigning lower uncertainty to more representative parts and higher uncertainty to less representative ones for the whole scene. This representativeness is then incorporated into the contrastive objective with uncertainty-guided weights. Finally, the uncertainty is further calibrated with an entailment loss regularized by entropy-based term. With the proposed losses, UNCHA learns hyperbolic embeddings with more accurate part-whole ordering, capturing the underlying compositional structure in an image and improving its understanding of complex multi-object scenes. UNCHA achieves state-of-the-art performance on zero-shot classification, retrieval, and multi-label classification benchmarks. Our code and models are available at: \url{https://github.com/jeeit17/UNCHA.git}.
\end{abstract}


\section{Introduction}
\label{sec:intro}
Understanding hierarchical structures is essential for capturing complex compositional information efficiently. As well established in cognitive science, human perception relies on part-whole hierarchies~\cite{HINTON1979231, hinton2023represent}, enabling generalization by interpreting new inputs through known relational structures~\cite{hinton2023represent, ito2022compositional, whittington2018generalisation}. Such hierarchical representations also improve information compression, classification, and inference efficiency~\cite{wu2022learning, chen2020towards, nickel2017poincare, ganea2018hyperbolic}. Vision-Language Models (VLMs) such as CLIP~\cite{radford2021learning}, ALIGN~\cite{jia2021scaling}, and ALBEF~\cite{li2021align} have demonstrated remarkable performance in image-text matching and shown strong versatility across various downstream tasks. However, owing to their reliance on Euclidean geometry, these models often face distortion of hierarchical structure and dimensionality trade-offs in capturing hierarchical or complex relational structures~\cite{he2025position, vendrov2015order, nickel2017poincare}. Moreover, CLIP has been reported to exhibit bias and difficulty with compositional relations in complex multi-object scenes~\cite{abbasi2025clip}, which is partly due to the lack of modeling part-whole relations. 

Hyperbolic space, characterized by constant negative curvature and exponential volume growth, provides an efficient geometric foundation for embedding hierarchical and fine-grained relational structures. Motivated by these properties, recent studies~\cite{khrulkov2020hyperbolicembeddings, chami2020trees, sala2018representation, dhingra2018embedding, pal2024compositional, ramasinghe2024accept, desai2023hyperbolic} have explored hyperbolic geometry in vision-language learning. MERU~\cite{desai2023hyperbolic} extended contrastive vision-language learning into hyperbolic space by explicitly modeling entailment relations between text and image pairs. ATMG~\cite{ramasinghe2024accept} later demonstrated that proximity-based contrastive losses can hinder hierarchical structure learning and proposed an angle-based alternative. HyCoCLIP~\cite{pal2024compositional} extended entailment modeling beyond inter-modal image-text relations by including intra-modal part-whole relationships. 

\begin{figure}[!t]
    \centering
     \includegraphics[width=0.5\textwidth]{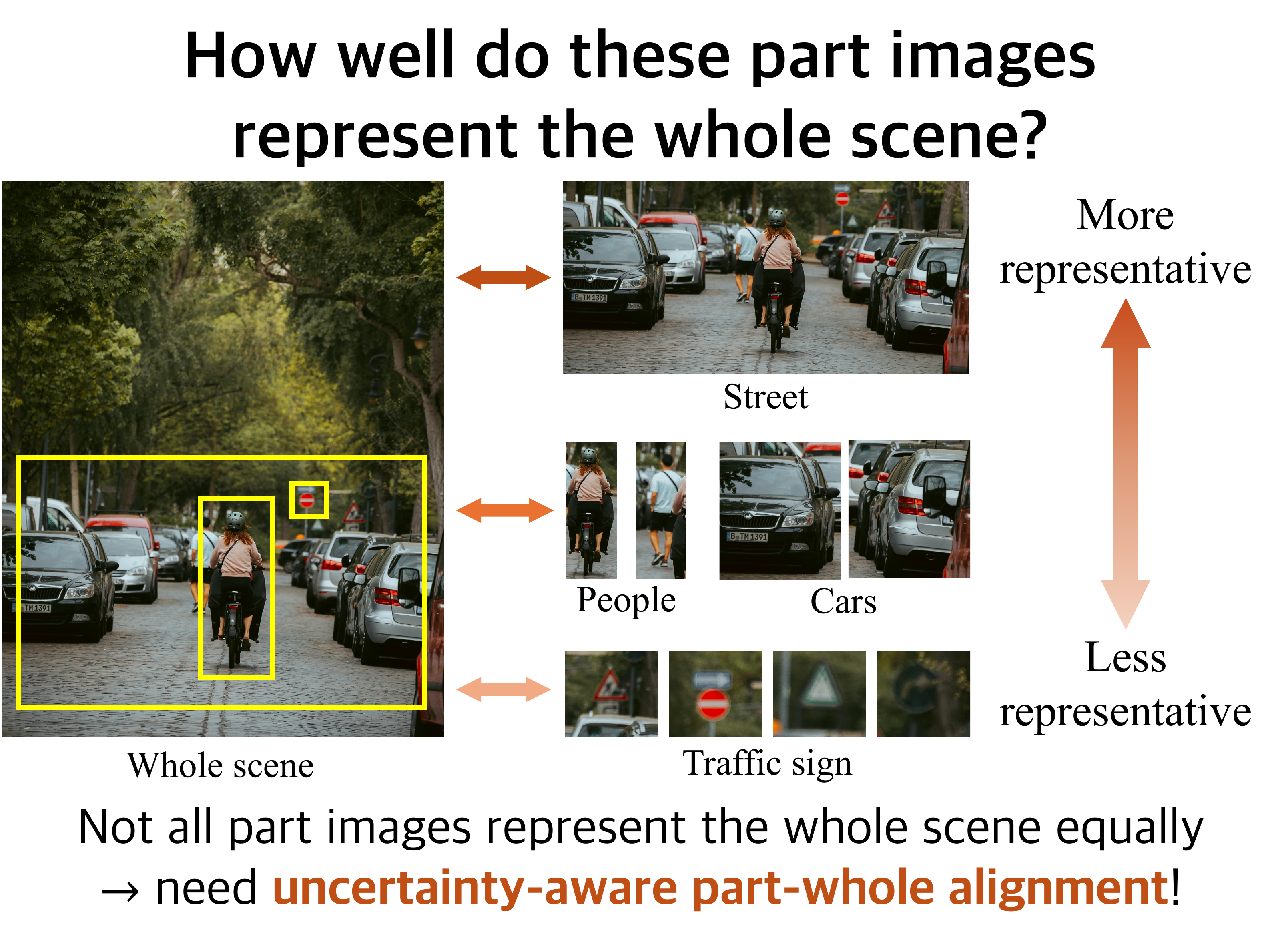}
     \vspace{-2em}
    \caption{\textbf{Varying representativeness of part images to whole scene.} The relationship between each part image and the whole scene varies with its representativeness. We model this varying representativeness as uncertainty, enabling uncertainty-guided part–whole alignment in hyperbolic space.}
    \label{fig:mainmoti}
   \vspace{-1em}
\end{figure}

Although hyperbolic approaches have demonstrated improved performance in hierarchy-aware representation learning, they do not model that each part has a different level of semantic representativeness to the whole. In other words, they do not account for the varying degree to which each part is semantically representative of the whole. As illustrated in Fig.~\ref{fig:mainmoti}, part images differ substantially in how well they represent the whole scene. When all parts are treated equally, the model may not appropriately distinguish more representative parts from less representative ones for the whole scene, often leading to degraded multi-object alignment and inefficient utilization of the embedding space~\cite{ramasinghe2024accept, pal2024compositional}.

We propose UNcertainty-guided Compositional Hyperbolic Alignment (UNCHA) for enhancing hyperbolic VLMs. UNCHA models part-to-whole semantic representativeness by assigning lower uncertainty to more representative parts and higher uncertainty to less representative ones for the whole scene. This design is grounded in prior findings~\cite{atigh2022hyperbolic, franco2023hyperbolic, yan2023hyp, mandica2024hyperbolic} showing that hyperbolic radius correlates with factors such as abstractness or uncertainty. Then, we incorporate uncertainty as part-to-whole semantic representativeness into both contrastive and entailment loss. Specifically, we incorporate uncertainty into the contrastive objective by assigning part-dependent temperature or uncertainty-guided weights, thereby modulating the strength of each part’s alignment with the whole. For the entailment loss, uncertainty is further calibrated based on the degree of part-to-whole entailment, and the entropy-based regularizer is also adapted to stabilize uncertainty estimates and promote richer use of the embedding space.
By continually training with the proposed losses, UNCHA progressively strengthens the semantic relationship across parts and wholes, leading to more accurate part-whole ordering in hyperbolic embeddings, capturing the underlying compositional structure in an image and improving its understanding of complex multi-object scenes.

We demonstrated that UNCHA outperforms prior hyperbolic VLMs~\cite{pal2024compositional, ramasinghe2024accept, desai2023hyperbolic} in diverse downstream tasks such as zero-shot image classification, retrieval, and a range of compositional and multi-object benchmarks, validating UNCHA's modeling of part-to-whole semantic representativeness and capability of more faithful compositional understanding. Our embedding space analysis further confirms UNCHA’s more discriminative and efficient use of part-to-whole modeling.
The contributions of this work are summarized as:
\begin{itemize}
\item We propose UNCHA, a uncertainty-guided compositional alignment with part-to-whole semantic representativeness, enabling hierarchy-aware and compositional representation learning for hyperbolic VLMs.
\item We model part-to-whole semantic representativeness with hyperbolic uncertainty, designing uncertainty-guided contrastive and entailment loss for uncertainty calibration, regularized by entropy to adaptively reflect part–whole relations.
\item We performed diverse benchmarks, demonstrating that UNCHA achieves superior performance over prior arts in downstream tasks such as retrieval, zero-shot and multi-object classification, validating the effectiveness of our uncertainty-guided compositional alignment.
\end{itemize}


\section{Related Works}

\subsection{Vision-language models}
Vision-Language Models (VLMs) have demonstrated strong capability in aligning image and text representations within a shared semantic space, achieving remarkable performance across tasks such as image-text retrieval and zero-shot image classification. The foundations of these models trace back to early studies on vision-language representation learning such as image retrieval, image captioning, and visual grounding, where joint embedding spaces are learned under task-specific supervision to associate visual content with linguistic semantics~\cite{lu2019vilbert, huang2020pixel, he2017fine, kiros2014unifying, ren2016joint, wu2019unified}. More recently, CLIP~\cite{radford2021learning} introduced a contrastive objective for aligning the two modalities using paired image-text data, achieving strong zero-shot and cross-modal performance~\cite{ge2023improving, pratt2023does, sain2023clip, jin2023refclip, sarkar2025crossover}. ALIGN~\cite{jia2021scaling} and ALBEF~\cite{li2021align} further extend CLIP by scaling up weak supervision and incorporating enhanced alignment-fusion strategies to better exploit large-scale, noisy datasets. 

However, the inherent limitations of Euclidean space make it difficult to represent hierarchical relationships effectively~\cite{nickel2017poincare, ibrahimi2024intriguing, peng2021hyperbolic}. Moreover, CLIP has been shown to exhibit biases in complex multi-object scenes~\cite{abbasi2025clip}. Its text encoder tends to emphasize the object mentioned first in the caption, while its image encoder focuses on larger objects, which hinders performance in multi-object settings. In contrast, hyperbolic space naturally provides continuous tree-like structures that support hierarchical embedding. However, when hierarchical relationships are handled without distinguishing their varying different part-to-whole representativeness, the embeddings tend to lose meaningful structural separation and collapse toward a narrow region~\cite{ramasinghe2024accept, pal2024compositional}. To address this, we introduce a part-to-whole uncertainty-guided alignment framework and explicitly model diverse part-whole entailment relationships within and across modalities, thereby enhancing compositional understanding.

\subsection{Hyperbolic representation learning} 
Hyperbolic space has emerged as an intriguing alternative in representation learning for embedding hierarchies. Hyperbolic space has exponential volume growth and a tree-like geometry, enabling near distortion-free hierarchical embeddings~\cite{ganea2018hyperbolic, sala2018representation}. Therefore, it provides an efficient representation for hierarchical structures. Consequently, numerous studies have leveraged hyperbolic geometry for representing text~\cite{tifrea2018poincar, dhingra2018embedding, le2019inferring}, images~\cite{khrulkov2020hyperbolicembeddings, yan2023hyp, atigh2022hyperbolic}, and graphs~\cite{liu2019hyperbolic, chami2019hyperbolic, sinha2024learning}. Recently, hyperbolic space has been integrated into foundation models to better capture hierarchical, compositional, and multi-modal structures at scale, enabling more expressive representations~\cite{he2025hyperbolic, desai2023hyperbolic, ramasinghe2024accept, pal2024compositional, he2025hypercore, mandica2024hyperbolic}. MERU~\cite{desai2023hyperbolic} first introduced hyperbolic vision–language models by employing an additional entailment loss~\cite{ganea2018hyperbolic, le2019inferring} inspired by order embeddings~\cite{vendrov2015order} to reflect the informativeness of different modalities. ATMG~\cite{ramasinghe2024accept} addressed hierarchical distortion and modality gap caused by spatial proximity–based contrastive learning by introducing an angle-based metric for image-text alignment in hyperbolic. HyCoCLIP~\cite{pal2024compositional} further incorporated intra-modal relationships by considering box images and their corresponding texts.

However, it does not differentiate the varying strengths of these relationships, resulting in limited distinction among parts. Several studies have explored the use of hyperbolic radius, the distance between an embedding and the origin, as a proxy for concept abstractness or uncertainty~\cite{atigh2022hyperbolic, franco2023hyperbolic, yan2023hyp, mandica2024hyperbolic}. The hyperbolic radius naturally provides uncertainty estimation and boundary awareness in pixel-level classification~\cite{atigh2022hyperbolic, franco2023hyperbolic}, image retrieval~\cite{yan2023hyp}, and multi-modal language understanding~\cite{mandica2024hyperbolic}, where it serves as an implicit indicator of confidence. Building on this property, we leverage the hyperbolic radius to better encode hierarchical structures in VLM and utilize entailment relationships for effective uncertainty calibration. An entropy-based regularizer further stabilizes the calibrated uncertainty, enabling more efficient use of the embedding space.


\section{Method}
\begin{figure*}[!t]
    \centering
   \vspace{-0.8em}
     \includegraphics[width=0.93\textwidth]{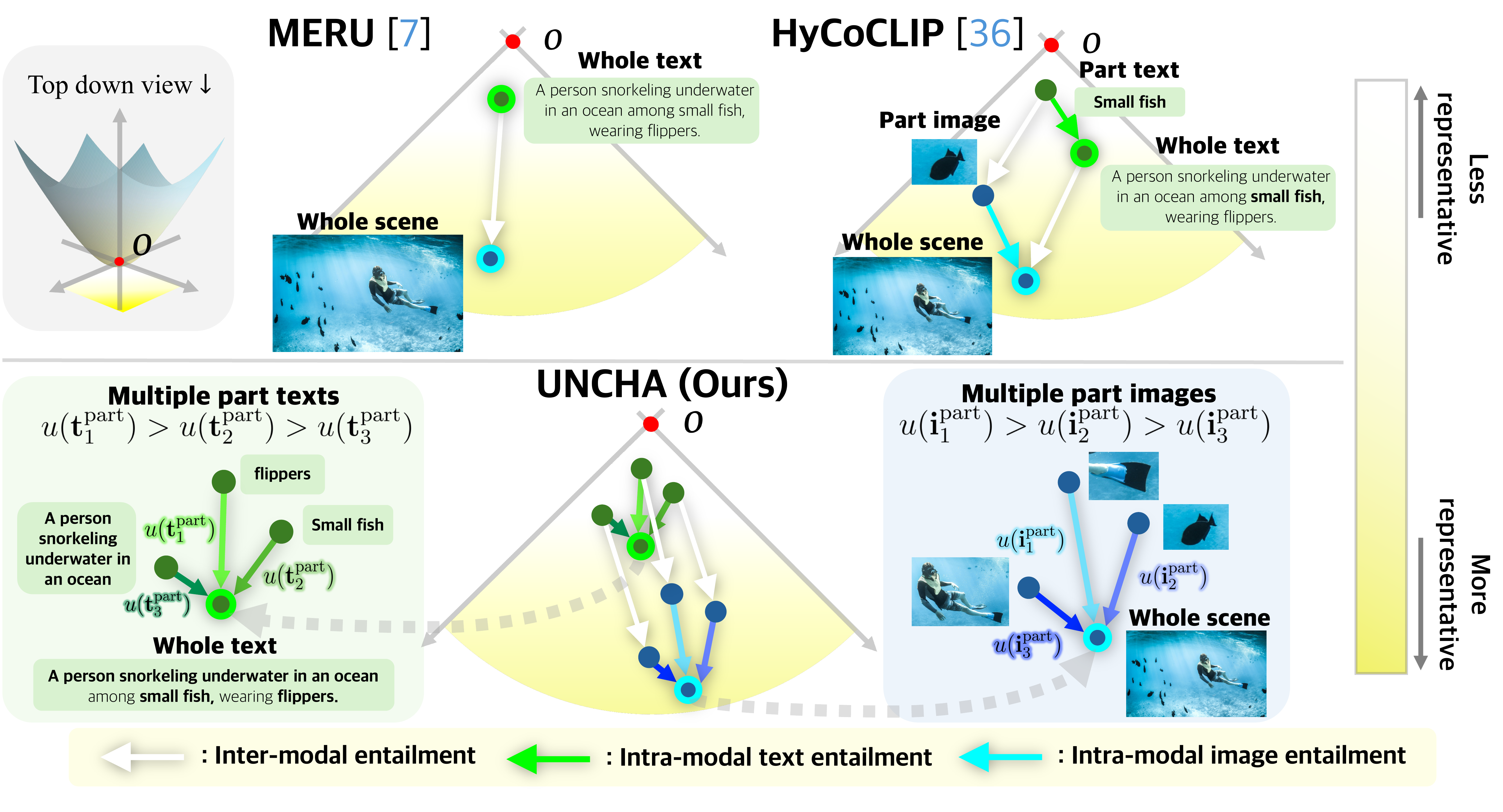}
     \vspace{-1em}
    \caption{\textbf{Comparison of UNcertainty-guided Compositional Hyperbolic Alignment (UNCHA, Ours) with prior works.} MERU~\cite{desai2023hyperbolic} models inter-modal entailment between whole scene image and text representations. HyCoCLIP~\cite{pal2024compositional} extends this to include intra-modal entailment between part and whole scene representations. UNCHA (Ours) further incorporates \emph{uncertainty to quantify the semantic representativeness} of each part, enabling uncertainty-guided part–whole alignment via adaptive weighting in the contrastive objectives and uncertainty calibration through the entailment loss. In addition, entropy regularization is applied in uncertainty calibration to ensure consistent and balanced utilization of the hyperbolic embedding space across varying uncertainty levels and modalities.}
    \label{fig:method}
       \vspace{-0.8em}
    \setlength{\intextsep}{5pt} 
\end{figure*}

\subsection{Preliminaries}

Hyperbolic space is a non-Euclidean geometry with a constant negative curvature $-\kappa$ where $\kappa \in \mathbb{R}^+$. Among several equivalent models, we adopt the Lorentz (or hyperboloid) model for embedding. A vector $\textbf{p} \in \mathbb{R}^{n+1}$ can be expressed in the form $[ p_{\text{time}},\mathbf{p}_{\text{space}}]$, where $\mathbf{p}_{\text{space}} \in \mathbb{R}^n$ and $p_{\text{time}} \in \mathbb{R}$. 
The Lorentzian inner product between two vectors $\mathbf{p}, \mathbf{q} \in \mathbb{R}^{n+1}$ is defined as:
\begin{equation}
\langle \mathbf{p}, \mathbf{q} \rangle_{\mathbb{L}} = -p_{\text{time}} q_{\text{time}} + \langle \mathbf{p}_{\text{space}}, \mathbf{q}_{\text{space}} \rangle,
\end{equation}
where $\langle \cdot , \cdot \rangle$ denotes the Euclidean inner product. 
The $n$-dimensional Lorentz manifold $\mathbb{L}^n$ is defined as the upper sheet of a two-sheeted hyperboloid in $(n+1)$-dimensional Minkowski space:
\begin{equation}
\mathbb{L}^n = \left\{ \mathbf{p} \in \mathbb{R}^{n+1} \; | \; 
\langle \mathbf{p}, \mathbf{p} \rangle_{\mathbb{L}} = -\frac{1}{\kappa}, \; \kappa > 0 \right\}.
\end{equation}
The geodesic distance between two points $\mathbf{p}, \mathbf{q}$ on the $n$-dimensional Lorentz manifold $\mathbb{L}^n$ is:
\begin{equation}
d_{\mathbb{L}}(\mathbf{p}, \mathbf{q}) = \sqrt{1/\kappa} \, \cosh^{-1}\left(-\kappa \langle \mathbf{p}, \mathbf{q} \rangle_{\mathbb{L}}\right).
\end{equation}
The hyperbolic radius of the embedding  $\mathbf{p}$ is defined as the geodesic distance from the origin of the hyperboloid $\mathbf{o}$, \textit{i.e.,} $d_{\mathbb{L}}(\mathbf{p}, \mathbf{o})$.

The tangent space at the point $\mathbf{z} \in \mathbb{L}^n$ is defined as:
\begin{equation}
T_{\mathbf{z}}\mathbb{L}^n = \left\{ \mathbf{v} \in \mathbb{R}^{n+1} : \langle \mathbf{z}, \mathbf{v} \rangle_{\mathbb{L}} = 0 \right\},
\end{equation}
which consists of Euclidean vectors $\mathbf{v}$ orthogonal to $\mathbf{z}$ under the Lorentzian inner product.
The exponential map projects a tangent vector $\mathbf{v} \in T_{\mathbf{z}}\mathbb{L}^n$ onto the manifold as below:
\begin{equation}
\exp^{\kappa}_{\mathbf{z}}(\mathbf{v}) =
\cosh(\sqrt{\kappa} \, \| \mathbf{v} \|_{\mathbb{L}})\mathbf{z} +
\frac{\sinh(\sqrt{\kappa} \, \| \mathbf{v} \|_{\mathbb{L}})}{\sqrt{\kappa} \, \| \mathbf{v} \|_{\mathbb{L}}}\mathbf{v}.
\end{equation}
Conversely, the logarithmic map sends a point $\mathbf{p} \in \mathbb{L}^n$ back to the tangent space at $\mathbf{z}$ as below:
\begin{equation}
\log^{\kappa}_{\mathbf{z}}(\mathbf{p}) =
\frac{\cosh^{-1}(-\kappa \langle \mathbf{z}, \mathbf{p} \rangle_{\mathbb{L}})}%
{\sqrt{(\kappa \langle \mathbf{z}, \mathbf{p} \rangle_{\mathbb{L}})^2 - 1}}
\; \mathrm{proj}_{\mathbf{z}}(\mathbf{p})
\end{equation}
where $\mathrm{proj}_{\mathbf{z}}(\mathbf{p}) = \mathbf{p} + \kappa \, \langle \mathbf{z}, \mathbf{p} \rangle_{\mathbb{L}} \mathbf{z}$. 
Here, we consider the case where $\mathbf{z}$ corresponds to the origin of the hyperboloid, $\mathbf{o} = [\sqrt{1/\kappa}, \mathbf{0}]$. In this setting, the time component of vectors in the tangent Euclidean space can be treated as zero, allowing us to parameterize the space component only, which is consistent with the design of prior works~\cite{desai2023hyperbolic, pal2024compositional, ramasinghe2024accept}. 

\subsection{Uncertainty-guided hyperbolic alignment}
\paragraph{Revisiting prior arts in hyperbolic alignment.}
Prior hyperbolic VLMs~\cite{ramasinghe2024accept, desai2023hyperbolic, pal2024compositional} extend contrastive vision-language learning by defining entailment relationships. In this hyperbolic geometry, abstract concepts tend to lie closer to the origin and specific ones farther out, with each specific concept constrained to its parent's entailment cone (see Sec~\ref{subsubsec:entailment} for details). As illustrated in Fig.~\ref{fig:method}, MERU~\cite{desai2023hyperbolic} incorporates an image-text entailment objective following partial-order embeddings~\cite{vendrov2015order}, where text is considered more abstract than image. HyCoCLIP~\cite{pal2024compositional} extends this idea by modeling intra-modal alignment, assuming that part image is more abstract than its corresponding whole scene. 
\paragraph{Method overview.}

\subsubsection{Uncertainty model of semantic representativeness}
 We leverage the geodesic distance from the origin (radius) in hyperbolic space~\cite{atigh2022hyperbolic, franco2023hyperbolic, yan2023hyp, mandica2024hyperbolic} to quantify the part-to-whole semantic representativeness using hyperbolic uncertainty. Since more abstract concepts are typically located near the origin and more specific ones farther away, this measure naturally reflects representativeness. Thus, we design the hyperbolic uncertainty to assign lower uncertainty to parts that are more representative of the whole scene, and high uncertainty otherwise ($e.g.,$ part images). As shown in Fig.~\ref{fig:analysis_sr}, our estimated uncertainty well aligns with semantic representativeness, indicating that the model effectively captures the varying part-to-whole relationships. 
 
Specifically, for a point $\mathbf{x} \in \mathbb{L}^n$, the Euclidean norm of $\mathbf{x}$ is monotonically related to its hyperbolic radius (see the supplementary material Sec.~S.2.3.1). Accordingly, we define the uncertainty $u$ as follows: 
\begin{equation}
u(\mathbf{x}) = \log\!\left(1 + \exp\!\left(-\|\mathbf{x}\|_2\right)\right).
\label{eq:hyperbolicuncertainty}
\end{equation}
Since points near the origin correspond to higher semantic uncertainty, the hyperbolic radius is inversely monotonically related to uncertainty. Eq.~\ref{eq:hyperbolicuncertainty} is a smooth monotonic transformation of the hyperbolic radius, which is a differentiable, well-behaved uncertainty measure for numerical stability.

\subsubsection{Uncertainty-guided contrastive loss}
In image–text pretraining, contrastive objectives are commonly employed to align multi-modal representations. Following prior works~\cite{desai2023hyperbolic, pal2024compositional}, we adopt the negative Lorentzian distance as the similarity measure as below:
\begin{equation}
L_{\text{c}}^{*}(\textbf{i}, \textbf{t}; \tau) = - \sum_i \log 
  \frac{
      \exp \left( -d_{\mathbb{L}}(\textbf{i}_i, \textbf{t}_i) / \tau \right)
  }{
      \sum_{\substack{k \neq i}} \exp \left( -d_{\mathbb{L}}(\textbf{i}_i, \textbf{t}_k) / \tau \right)
  }
\label{eq:contrastive_basic}
\end{equation} 
where the $i$-th image embedding $\textbf{i}_i$ and its corresponding text embedding $\textbf{t}_i$ form a \textit{positive} pair while all other text embeddings $\textbf{t}_i$ with $k \neq i$ are treated as \textit{negatives} in the batch of size $B$ and the temperature parameter $\tau$ controls the scaling of similarities. 

Prior work~\cite{pal2024compositional} introduces a global–local contrastive loss $\mathcal{L}_{\text{con}}^{\text{orig}}$ that aligns part-level text features $\mathbf{t}^{\text{part}}$ with whole image embeddings, and part-level image features $\mathbf{i}^{\text{part}}$ with whole text embeddings as below: 
\begin{equation}
  \underbrace{L_{\text{c}}^{*}\!\left(\mathbf{i}^{\text{part}}, \mathbf{t}; \tau \right)
   + L_{\text{c}}^{*}\!\left(\mathbf{t}^{\text{part}}, \mathbf{i}; \tau\right)}_{\text{global-local contrastive loss}} 
   + \underbrace{L_{\text{c}}^{*}(\mathbf{i}, \mathbf{t}; \tau)
   + L_{\text{c}}^{*}(\mathbf{t}, \mathbf{i}; \tau)}_{\text{global contrastive loss}}.
\label{eq:prior_contrasitve_loss}
\end{equation}

Our contrastive loss additionally includes a local contrastive loss that explicitly aligns each part image with its corresponding text on top of Eq.~\ref{eq:prior_contrasitve_loss}. Since whole and part images differ in information levels and occupy distinct regions in hyperbolic space, we design to assign separate temperature parameters, $\tau_{g}$, $\tau_{l}$, and $\tau_{gl}$ to global, local and global-local contrastive losses, respectively, to better model these relationships. 

We propose uncertainty-guided contrastive loss unlike the aforementioned prior contrastive losses with fixed temperature.
Our approach incorporates uncertainty into the global-local contrastive loss by considering the varying semantic representativeness of multiple parts.  We modulate the temperature in an element-wise manner through an uncertainty-guided global-local contrastive loss, where the temperature is adaptively scaled according to the estimated uncertainty of each part image and text. The adaptive temperatures $\boldsymbol{\tau}_{\text{un},i}^{I}$ and $\boldsymbol{\tau}_{\text{un},i}^{T}$ are designed as below:

\begin{equation}
\resizebox{0.9\linewidth}{!}{$
\boldsymbol{\tau}^{I}_{\text{un}, i}
= \exp\!\left(u(\mathbf{i}^{\text{part}}_{i})/2\right)\,\tau_{gl},
\boldsymbol{\tau}^{T}_{\text{un}, i}
= \exp\!\left(u(\mathbf{t}^{\text{part}}_{i})/2\right)\,\tau_{gl}
$}
\label{eq:uncertainty_scaled_temp}
\end{equation}
where higher uncertainty leads to a larger temperature and a smaller contribution to the contrastive loss.
The formulation of our proposed contrastive loss is shown as below: 
\begin{eqnarray}
\mathcal{L}_{\text{con}}^{\text{un}}
  &=& 
  \underbrace{
  L_{\text{c}}^{*}\!\left(\mathbf{i}^{\text{part}}, \mathbf{t}; \boldsymbol{\tau}_{\text{un}}^{I}\right)
  + L_{\text{c}}^{*}\!\left(\mathbf{t}^{\text{part}}, \mathbf{i}; \boldsymbol{\tau}_{\text{un}}^{T}\right)
  }_{\substack{\text{uncertainty-guided global-local contrastive loss}}} \label{eq:ours_contrastive_loss} \\
  &&+ \underbrace{
  L_{\text{c}}^{*}(\mathbf{i}, \mathbf{t}; \tau_{g})
  + L_{\text{c}}^{*}(\mathbf{t}, \mathbf{i}; \tau_{g})
  }_{\substack{\text{global contrastive loss}}} \nonumber \\
  &&+ 
  \underbrace{
  L_{\text{c}}^{*}(\mathbf{i}^{\text{part}}, \mathbf{t}^{\text{part}}; \tau_{l})
  + L_{\text{c}}^{*}(\mathbf{t}^{\text{part}}, \mathbf{i}^{\text{part}}; \tau_{l})
  }_{\substack{\text{local contrastive loss}}}. \nonumber
\end{eqnarray}

Unlike the one-to-one correspondence between matched image-text pairs, the relationship between a part image and its whole scene or text may not be a perfect correspondence. For instance, a single scene text may correspond to multiple part images. If all embeddings within a whole scene are pushed apart with the same temperature, both highly representative and less representative regions are equally repelled, breaking semantic structure. Our proposed contrastive loss in Eq.~\ref{eq:ours_contrastive_loss} is designed to mitigate these undesirable cases.

\begin{figure}[!t]
    \centering
    \includegraphics[width=0.42\textwidth]{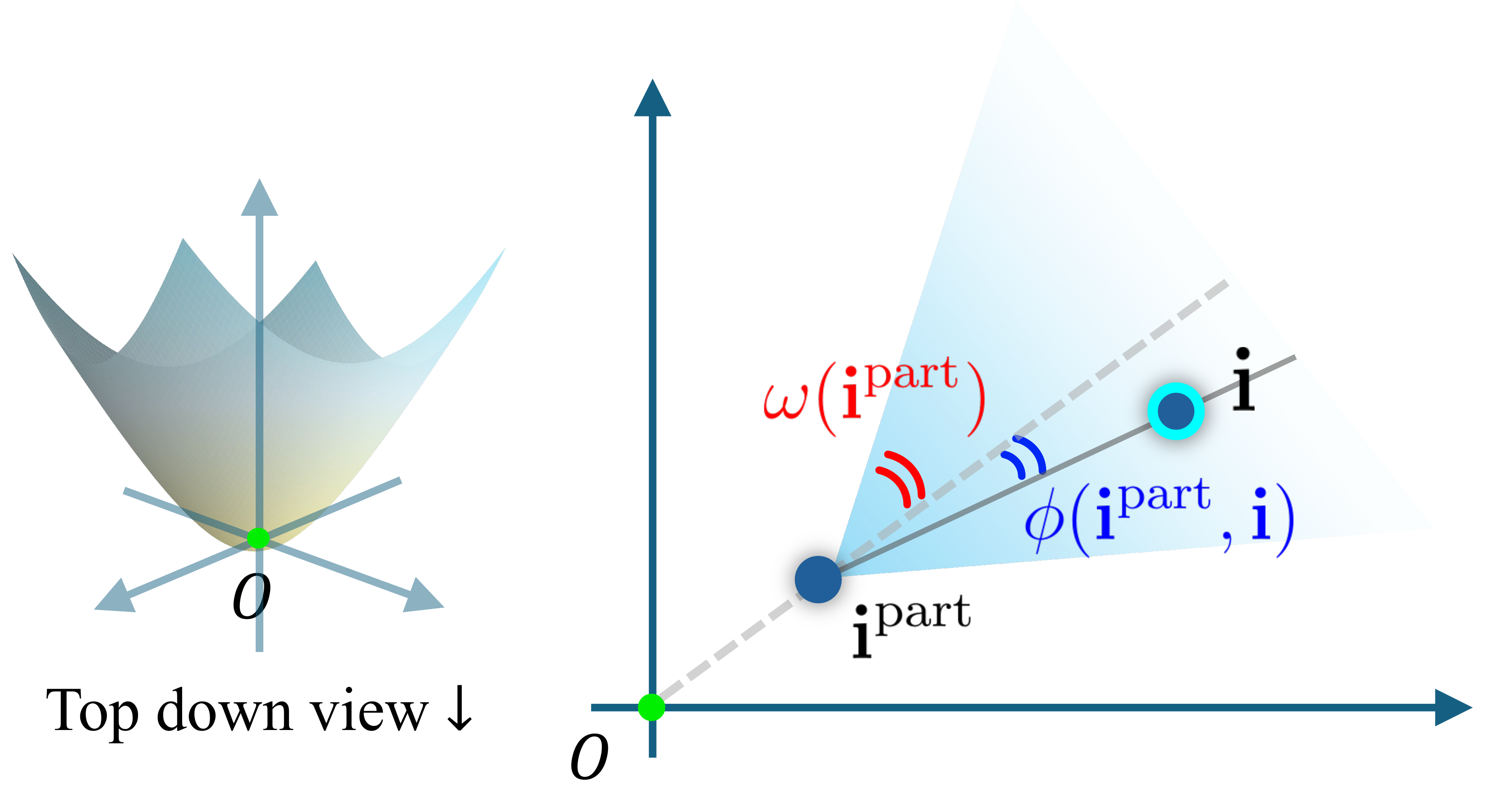} 
    \vspace{-0.8em}
    \caption{\textbf{Entailment geometry in hyperbolic space.} 
    The term $\omega(\mathbf{i}^{\text{part}})$ denotes the aperture of the entailment cone centered at $\mathbf{i}^{\text{part}}$. 
    The angle $\phi(\mathbf{i}^{\text{part}}, \mathbf{i})$ measures the geodesic angle between the embeddings $\mathbf{i}^{\text{part}}$ and $\mathbf{i}$, which is used to determine whether $\mathbf{i}$ lies within the entailment region of $\mathbf{i}^{\text{part}}$.}
    \label{fig:entailment}
\vspace{-1em}
\end{figure}

\subsubsection{Entailment loss for uncertainty calibration}
\label{subsubsec:entailment}
\paragraph{Piecewise-continuous entailment loss.} Building upon the hyperbolic entailment formulation in~\cite{desai2023hyperbolic, le2019inferring}, prior work~\cite{pal2024compositional} defines the entailment loss as:
\begin{equation}
\mathcal{L}_{\text{orig}}= \max(0, \phi(\mathbf{p}, \mathbf{q}) - \eta  \omega(\mathbf{p}))
\label{eq:entailloss}
\end{equation}
where $\phi(\mathbf{p}, \mathbf{q})$ denotes the angular distance between the embeddings $\mathbf{p}$ and $\mathbf{q}$, 
$\eta$ and $K$ are hyperparameters, and $\omega(\mathbf{p})$ defines the aperture of the entailment cone centered at $\mathbf{p}$ as below:
\begin{equation}
\omega(\mathbf{p}) = \sin^{-1}\!\left( {2K}/{(\sqrt{-\kappa}\|\mathbf{p}\| )} \right),
\end{equation}
which is also illustrated in Fig.~\ref{fig:entailment}. The $\mathcal{L}_{\text{orig}}$ in Eq.~\ref{eq:entailloss} enforces entailment by constraining $\mathbf{q}$ to lie within the cone of $\mathbf{p}$. 
However, once $\mathbf{q}$ is fully contained in the cone, the loss becomes zero, preventing further fine-grained alignment. 

Here, we propose adding an angular term $\phi(\mathbf{p}, \mathbf{q})$ in Eq.~\ref{eq:entailloss} to encourage fine-grained alignment while maintaining smooth optimization continuity as below:
\begin{equation}
L^{*}_{\text{ent}}(\mathbf{p}, \mathbf{q})
  = \max\!\left(0,\, \phi(\mathbf{p}, \mathbf{q}) - \eta\,\omega(\mathbf{p})\right)
  + \alpha\,\phi(\mathbf{p}, \mathbf{q})
\label{eq:piecewise_continuous}
\end{equation}
where $\alpha$ is a hyperparameter. This formulation can be viewed as a Leaky-ReLU-like~\cite{maas2013rectifier} relaxation of the original hinge-based entailment loss, with the additional term preserving a small gradient even when $\mathbf{q}$ is inside the cone.

\paragraph{Uncertainty calibration loss.} Prior studies have reported that hyperbolic embeddings often accumulate around narrow regions, leading to collapse~\cite{ramasinghe2024accept}. Moreover, local and global image representations exhibit similar radii, making their separation less distinct~\cite{pal2024compositional}. To clearly distinct global and local representations, we propose the uncertainty calibration loss as follows: 
\begin{equation}
L_{\text{ent}}^{\text{cal}}(\mathbf{p}, \mathbf{q})
  = \left\lfloor L_{\text{ent}}^{*}(\mathbf{p}, \mathbf{q}) \right\rfloor \,
    \! e^{-u(\mathbf{p})} + u(\mathbf{p}) +  \mathcal{H}(\tilde{u}({\mathbf{p}}))
\label{eq:cal2}
\end{equation}
where $\left\lfloor \, . \right\rfloor$ denotes the stop-gradient operator and $\mathcal{H}$ represents the entropy term as follows:
\begin{equation}
    \mathcal{H}(\tilde{u}({\mathbf{p}})) = - \sum\nolimits_i \tilde{u}({\mathbf{p}_i}) \log(\tilde{u}({\mathbf{p}_i}))
\label{eq:cal1}
\end{equation}
where \( \tilde{u}({\mathbf{p}_i}) = {\exp(u(\mathbf{p}_i))} / {\sum_j \exp(u(\mathbf{p}_j))} \).
When the entailment relation between $\mathbf{p}$ and $\mathbf{q}$ is weak, the term $e^{-u(\mathbf{p})}$ encourages the model to increase uncertainty. The term $u(\mathbf{p})$ prevents the model from assigning excessively high uncertainty just to reduce the loss. Thus, $\mathcal{H}(\tilde{u}({\mathbf{p}}))$ regularizes the uncertainty distribution to remain diverse and informative, avoiding a collapse toward uniform or constant uncertainty, analogous to~\cite{grandvalet2004semi}.

\begin{figure}[!t]
    \centering
    \includegraphics[width=0.42\textwidth]{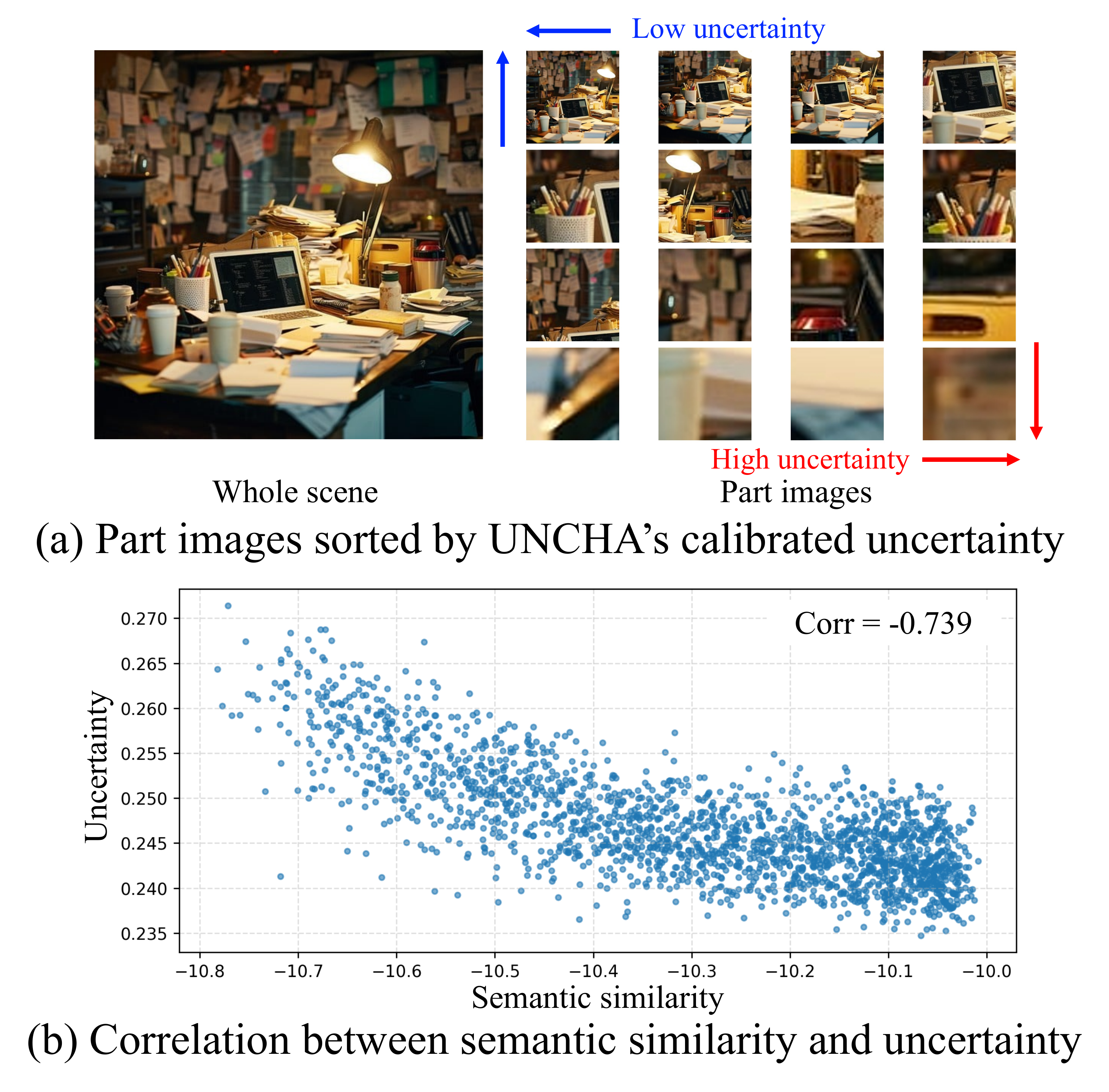} 
    \vspace{-0.8em}
    \caption{\textbf{Analysis of uncertainty modeling.} 
    (a) Randomly cropped parts are sorted by uncertainty (low→high). Semantically representative parts show low uncertainty, while blurred or less representative crops show high uncertainty. 
    (b) On an ImageNet~\cite{russakovsky2015imagenet} subset, part-to-whole similarity vs. uncertainty shows a strong negative correlation ($r = -0.739$), indicating that less representative parts have higher uncertainty.}
    \label{fig:analysis_sr}
   \vspace{-1em}
\end{figure}

With the entropy regularizer, the proposed formulation of our entailment loss is as follows:
\begin{eqnarray}
\label{eq:uncertainty_total}
\mathcal{L}_{\text{ent}}^{\text{un}} &=&
\underbrace{
L_{\text{ent}}^{*}(\mathbf{t}^{\text{part}}, \mathbf{i}^{\text{part}})
+ L_{\text{ent}}^{*}(\mathbf{t}, \mathbf{i})
}_{\text{inter-modal entailment}}  \\
&&+ \lambda_{1} (\underbrace{L_{\text{ent}}^{*}(\mathbf{t}^{\text{part}}, \mathbf{t})
+ L_{\text{ent}}^{*}(\mathbf{i}^{\text{part}}, \mathbf{i})}_{\text{intra-modal entailment}} \nonumber )\\
&&+ \lambda_{2}(\underbrace{
L_{\text{ent}}^{cal}(\mathbf{t}^{\text{part}}, \mathbf{t})
 + L_{\text{ent}}^{cal}(\mathbf{i}^{\text{part}}, \mathbf{i}))}_{\text{uncertainty calibration}} 
\nonumber
\end{eqnarray}
where $\lambda_{1}$ and $\lambda_{2}$ are hyperparameters. This uncertainty calibration enables semantic alignment with the representativeness of each part relative to the whole. This is a process that naturally fits the geometric properties of hyperbolic space, which is particularly beneficial for jointly aligning multiple objects simultaneously. Moreover, such calibration enhances multi-object alignment, as shown in Fig.~\ref{fig:analysis_sr}. Parts with higher semantic similarity to the whole exhibit lower uncertainty, while less representative parts show higher uncertainty, resulting in a strong negative correlation between similarity and uncertainty. Further details on Fig.~\ref{fig:analysis_sr} are provided in the supplementary material Sec.~S.2.3.2.

Finally, our overall loss with the proposed uncertainty-guided contrastive loss in Eq.~\ref{eq:ours_contrastive_loss} and the entailment loss with uncertainty calibration in Eq.~\ref{eq:uncertainty_total} is 
defined as follows:
\begin{equation}
\begin{aligned}
L = \mathcal{L}_{\text{con}}^{\text{un}} + \lambda_{ent} \mathcal{L}_{\text{ent}}^{\text{un}} 
\end{aligned}
\end{equation}
where $\lambda_{ent}$ is a hyperparameter. We detail all hyperparameters in the supplementary material Sec.~S.1.2. 


\section{Experiments}

\begin{table*}[t]
\vspace{-0.5cm}
\centering
\small
\setlength{\tabcolsep}{3pt}
\renewcommand{\arraystretch}{1.2}

\setlength{\tabcolsep}{4.5pt} 
\renewcommand{\arraystretch}{0.95} 
\caption{\textbf{Zero-shot image classification evaluation.} UNCHA (Ours) consistently demonstrates strong zero-shot classification performance across both architectures. Bold numbers denote the best performance within each architecture. $\dagger$ denotes ATMG trained on the GRIT~\cite{peng2023kosmos}.}
\vspace{-1em}
\begin{tabular}{llcccccccccccccccc}

\toprule
& & \multicolumn{6}{c}{\cellcolor{lightblue}\textbf{General datasets}} & \multicolumn{6}{c}{\cellcolor{lightcyan}\textbf{Fine-grained datasets}} & \multicolumn{4}{c}{\cellcolor{lightgreen}\textbf{Misc. datasets}}\\
\cmidrule(lr){3-8} \cmidrule(lr){9-14} \cmidrule(lr){15-18}
 & \textbf{Model} & \rotatebox{90}{ImageNet} & \rotatebox{90}{CIFAR-10} & \rotatebox{90}{CIFAR-100} & \rotatebox{90}{SUN397} & \rotatebox{90}{Caltech-101} & \rotatebox{90}{STL-10} & \rotatebox{90}{Food-101} & \rotatebox{90}{CUB} & \rotatebox{90}{Cars} & \rotatebox{90}{Aircraft} & \rotatebox{90}{Pets} & \rotatebox{90}{Flowers} & \rotatebox{90}{DTD} & \rotatebox{90}{EuroSAT} & \rotatebox{90}{RESISC45} & \rotatebox{90}{Country211}\\
\midrule

\multirow{5}{*}{ViT-S/16}
& CLIP~\cite{radford2021learning} & 36.7 & 70.2 & 42.6 & 35.8 & 57.6 & 89.7 & 44.7 & 9.8 & 6.9 & 2.0 & 44.6 & 14.8 & 22.3 & \textbf{40.7} & 40.1 & 5.1 \\
& MERU~\cite{desai2023hyperbolic} & 35.4 & 71.2 & 40.4 & 33.8 & 57.3 & 89.7 & 41.2 & 11.3 & 5.2 & \textbf{4.2} & 42.7 & 17.3 & 18.6 & 39.1 & 38.9 & \textbf{5.3} \\
& ATMG$^\dagger$~\cite{ramasinghe2024accept} & 34.1 & 66.9 & 42.1 & 47.9 & 68.5 & 90.7 & 43.6 & 14.1 & 5.8 & 2.5 & 41.8 & 14.9 & 19.7 & 35.8 & 40.3 & 4.6 \\
& HyCoCLIP~\cite{pal2024compositional} & 41.7 & 85.0 & 53.4 & {{52.5}} & {75.7} & {92.5} & {50.2} & \textbf{{14.7}} & {8.1} & \textbf{4.2} & {52.0} & 20.5 & 23.3 & 38.3 & \textbf{{45.7}} & 5.2 \\
& \cellcolor{gray!10}\textbf{UNCHA (Ours)} 
& \cellcolor{gray!10}\textbf{43.9} & \cellcolor{gray!10}\textbf{85.9} & \cellcolor{gray!10}\textbf{56.6} 
& \cellcolor{gray!10}\textbf{52.6} & \cellcolor{gray!10}\textbf{80.5} & \cellcolor{gray!10}\textbf{94.4} 
& \cellcolor{gray!10}\textbf{52.1} & \cellcolor{gray!10}12.5
& \cellcolor{gray!10}\textbf{9.2} & \cellcolor{gray!10}2.7 
& \cellcolor{gray!10}\textbf{52.1} & \cellcolor{gray!10}\textbf{24.6}
& \cellcolor{gray!10}\textbf{25.4} & \cellcolor{gray!10}36.2
& \cellcolor{gray!10}43.4 & \cellcolor{gray!10}5.2 \\
\midrule

\multirow{5}{*}{ViT-B/16}
& CLIP~\cite{radford2021learning} & 40.6 & 78.9 & 48.3 & 43.0 & 70.7 & 92.4 & 48.3 & 10.4 & 9.3 & 3.4 & 45.9 & 21.3 & 23.4 & 37.1 & 42.7 & 5.7 \\
& MERU~\cite{desai2023hyperbolic}
 & 40.1 & 78.6 & 49.3 & 43.0 & 73.0 & 92.8 & 48.5 & 11.0 & 5.3 & 3.7 & 48.5 & 21.6 & 22.1 & 31.7 & 42.6 & 5.4 \\
& ATMG$^\dagger$~\cite{ramasinghe2024accept} & 34.3 & 68.8 & 42.1 & 48.2 & 68.5 & 91.2 & 43.2 & 14.3 & 6.0 & 2.4 & 42.2 & 15.0 & 19.4 & 35.0 & 40.4 & 4.6 \\
& HyCoCLIP~\cite{pal2024compositional} & 45.8 & 88.8 & 60.1 & 57.2 & 81.3 & 95.0 & 59.2 & \textbf{16.4} & 11.6 & 3.7 & 56.8 & 23.9 & 29.4 & 35.8 & 45.6 & \textbf{6.5} \\
& \cellcolor{gray!10}\textbf{UNCHA (Ours)} 
& \cellcolor{gray!10}\textbf{48.8} & \cellcolor{gray!10}\textbf{90.4} 
& \cellcolor{gray!10}\textbf{63.2} & \cellcolor{gray!10}\textbf{57.7} 
& \cellcolor{gray!10}\textbf{83.9} & \cellcolor{gray!10}\textbf{95.7} 
& \cellcolor{gray!10}\textbf{60.3} & \cellcolor{gray!10}14.8 
& \cellcolor{gray!10}\textbf{14.0} & \cellcolor{gray!10}\textbf{3.8} 
& \cellcolor{gray!10}\textbf{57.1} & \cellcolor{gray!10}\textbf{27.0}
& \cellcolor{gray!10}\textbf{30.3} & \cellcolor{gray!10}\textbf{41.3}
& \cellcolor{gray!10}\textbf{52.7} & \cellcolor{gray!10}6.1 \\
\bottomrule
\end{tabular}
\label{tab:zero_shot_cls}
\end{table*}

\begin{table*}[t]
\setlength{\abovecaptionskip}{4pt}
\setlength{\belowcaptionskip}{-6pt}
\renewcommand{\arraystretch}{0.92}

\centering
\small
\setlength{\tabcolsep}{4pt}

\caption{\textbf{Zero-shot retrieval and hierarchical classification metrics on ImageNet~\cite{deng2009imagenet}.} UNCHA (Ours) consistently achieves superior performance across both retrieval and hierarchical metrics, showing the effectiveness of our uncertainty-based hyperbolic alignment.}

\begin{tabular}{llccccccccccccccccccc}
\toprule
\multicolumn{3}{c}{} 
& \multicolumn{4}{c}{\cellcolor{lightblue}\textbf{Text retrieval}} 
& \multicolumn{4}{c}{\cellcolor{lightcyan}\textbf{Image retrieval}} 
& \multicolumn{5}{c}{\cellcolor{lightgreen}\textbf{Hierarchical metrics}} \\
\cmidrule(lr){4-7} \cmidrule(lr){8-11} \cmidrule(lr){12-16}

& \textbf{Model} 
&  & \multicolumn{2}{c}{\textbf{COCO}} & \multicolumn{2}{c}{\textbf{Flickr}} 
& \multicolumn{2}{c}{\textbf{COCO}} & \multicolumn{2}{c}{\textbf{Flickr}} 
& \multirow{2}{*}{\textbf{TIE(↓)}} 
& \multirow{2}{*}{\textbf{LCA(↓)}} 
& \multirow{2}{*}{\textbf{J(↑)}} 
& \multirow{2}{*}{$\mathbf{P_H}$(↑)} 
& \multirow{2}{*}{$\mathbf{R_H}$(↑)} \\

\cmidrule(lr){4-5} \cmidrule(lr){6-7}
\cmidrule(lr){8-9} \cmidrule(lr){10-11}

&  & 
& R@5 & R@10 & R@5 & R@10
& R@5 & R@10 & R@5 & R@10 
&  &  &  &  &  \\

\midrule

\multirow{5}{*}{ViT-S/16}
& CLIP~\cite{radford2021learning} &  & 69.3 & 79.1 & 90.2 & \textbf{95.2} & 53.7 & 65.2 & 81.1 & 87.9 
& 4.02 & 2.39 & 0.76 & 0.83 & 0.84 \\
& MERU~\cite{desai2023hyperbolic} &  & 68.8 & 78.8 & 89.4 & 94.8 & 53.6 & 65.3 & 80.4 & 87.5 
& 4.08 & 2.39 & 0.76 & 0.83 & 0.83 \\
& ATMG$^\dagger$~\cite{ramasinghe2024accept} &  & 62.6 & 74.2 & 85.5 & 91.6 & 50.3 & 62.1 & 76.9 & 84.6 
& 4.26 & 2.50 & 0.75 & 0.82 & 0.83 \\
& HyCoCLIP~\cite{pal2024compositional} &  & 69.5 & 79.5 & 89.1 & 93.9 & 55.2 & 66.6 & 81.5 & 88.1 
& 3.55 & 2.17 & 0.79 & \textbf{0.86} & 0.85 \\
& \cellcolor{gray!10}\textbf{UNCHA (Ours)} 
& \cellcolor{gray!10}  
& \cellcolor{gray!10} \textbf{69.9} & \cellcolor{gray!10}\textbf{79.7} 
& \cellcolor{gray!10}\textbf{90.8} & \cellcolor{gray!10}{94.8} 
& \cellcolor{gray!10}\textbf{56.2} & \cellcolor{gray!10}\textbf{67.6} 
& \cellcolor{gray!10}\textbf{82.5} & \cellcolor{gray!10}\textbf{89.3}
& \cellcolor{gray!10}\textbf{3.39} & \cellcolor{gray!10}\textbf{2.14} 
& \cellcolor{gray!10}\textbf{0.80} & \cellcolor{gray!10}\textbf{0.86}
& \cellcolor{gray!10}\textbf{0.86} \\
\midrule

\multirow{5}{*}{ViT-B/16}
& CLIP~\cite{radford2021learning} &  & 71.4 & 81.5 & \textbf{93.6} & \textbf{96.9} & 57.4 & 68.5 & 83.5 & 89.9 
& 3.60 & 2.21 & 0.79 & 0.85 & 0.85 \\
& MERU~\cite{desai2023hyperbolic} &  & 72.3 & 82.0 & 93.5 & 96.2 & 57.4 & 68.6 & 84.0 & 90.0 
& 3.63 & 2.22 & 0.78 & 0.85 & 0.85 \\
& ATMG$^\dagger$~\cite{ramasinghe2024accept} &  & 62.9 & 74.0 & 85.1 & 92.2 & 51.2 & 62.6 & 78.0 & 85.3 & 4.19 
& 2.48 & 0.75 & 0.83 & 0.83 \\
& HyCoCLIP~\cite{pal2024compositional} &  & 72.0 & 82.0 & 92.6 & 95.4 & 58.4 & 69.3 & \textbf{84.9} & 90.3 
& 3.17 & 2.05 & 0.81 & 0.87 & 0.87 \\
& \cellcolor{gray!10}\textbf{UNCHA (Ours)} 
& \cellcolor{gray!10} 
& \cellcolor{gray!10}\textbf{72.7} & \cellcolor{gray!10}\textbf{82.7} 
& \cellcolor{gray!10}91.4 & \cellcolor{gray!10}95.9 
& \cellcolor{gray!10}\textbf{60.0} & \cellcolor{gray!10}\textbf{71.0} 
& \cellcolor{gray!10}\textbf{84.9} & \cellcolor{gray!10}\textbf{91.2}
& \cellcolor{gray!10}\textbf{2.94} & \cellcolor{gray!10}\textbf{1.96} 
& \cellcolor{gray!10}\textbf{0.83} & \cellcolor{gray!10}\textbf{0.88}
& \cellcolor{gray!10}\textbf{0.88} \\
\bottomrule
\end{tabular}
\label{tab:retrieval_hier_imagenet}
\vspace{-1em}
\end{table*}

\begin{table}[t]
\centering
\small
\setlength{\tabcolsep}{7pt}
\renewcommand{\arraystretch}{1.05}
\caption{\textbf{Comparison on part-level alignment evaluation with hard negatives.} Ours achieves substantial performance gains under the  most challenging scenario of~\cite{Urbanek_2024_CVPR}, demonstrating its strong ability for fine-grained compositional understanding.}

\begin{tabular}{lcccc}
\toprule
& \multicolumn{2}{c}{\cellcolor{lightgreen}\textbf{All Pick5}}
& \multicolumn{1}{c}{\cellcolor{lightgreen}\textbf{All}} \\
\cmidrule(lr){2-3} \cmidrule(lr){4-4}

\textbf{Model}
& \textbf{SCM}
& \textbf{Neg}
& \textbf{Hard Negs} \\
\midrule

CLIP~\cite{radford2021learning}
& 13.10 & 22.94 & 52.89 \\

ATMG$^\dagger$~\cite{ramasinghe2024accept}
& 12.23 & 23.08 & 53.91 \\

MERU~\cite{desai2023hyperbolic}
& 12.59 & 20.69 & 54.56 \\

HyCoCLIP~\cite{pal2024compositional}
& 11.65 & 23.52 & 53.33 \\

\cellcolor{gray!10}\textbf{UNCHA (Ours)}
& \cellcolor{gray!10}\textbf{13.53}
& \cellcolor{gray!10}\textbf{23.81}
& \cellcolor{gray!10}\textbf{56.51} \\
\bottomrule
\end{tabular}
\vspace{-2em}
\label{tab:swap-hard}
\end{table}

\begin{table}[t]
\centering
\small
\setlength{\tabcolsep}{5pt}
\renewcommand{\arraystretch}{0.95}

\caption{\textbf{Ablation study on classification and retrieval benchmarks.} Removing any component leads to consistent performance drops, showing that all modules contribute meaningfully. Bold numbers indicate the best performance within each task group.}

\begin{tabular}{lcccccccc}
\toprule
& \multicolumn{3}{c}{\cellcolor{lightblue}\textbf{Classification}} 
& \multicolumn{2}{c}{\cellcolor{lightcyan}\textbf{Retrieval}} \\
\cmidrule(lr){2-4} \cmidrule(lr){5-6}
\textbf{Model} 
& \rotatebox{90}{General} 
& \rotatebox{90}{Fine} 
& \rotatebox{90}{MISC.} 
& \rotatebox{90}{Text} 
& \rotatebox{90}{Image} \\
\midrule

\cellcolor{gray!10}\textbf{Ours (full)} 
& \cellcolor{gray!10}\textbf{68.98} 
& \cellcolor{gray!10}\textbf{25.53} 
& \cellcolor{gray!10}\textbf{27.55} 
& \cellcolor{gray!10}\textbf{83.80} 
& \cellcolor{gray!10}\textbf{73.90} \\

w/o uncertainty 
& 64.57 
& 22.98 
& 26.67 
& 79.60 
& 69.68 \\

w/o contrastive 
& 65.14 
& 23.92 
& 25.58 
& 80.78 
& 70.55 \\

w/o entropy
& 65.61 
& 23.09 
& 24.78 
& 80.60 
& 69.95 \\
\bottomrule
\end{tabular}
\vspace{-2em}
\label{tab:ablation}
\end{table}

\begin{table*}[t]
\vspace{-0.6cm}
\centering
\small
\setlength{\tabcolsep}{5.3pt}  

\caption{\textbf{Comparison across Multi-object Representation and Classification tasks.}  
Left: zero-shot mAP comparison across multi-object configurations on ComCo and SimCo datasets.  
Right: zero-shot multi-label classification (Cls.) on VOC and COCO datasets (mAP only).  
Our method consistently achieves higher mAP across both tasks.}
\vspace{-0.5em}
\begin{tabular}{llcccccccccccc}
\toprule
\multicolumn{3}{c}{} 
& \multicolumn{8}{c}{\cellcolor{lightblue}\textbf{Multi-object Representation }} 
& \multicolumn{2}{c}{\cellcolor{lightgreen}\textbf{Multi-label Cls.}} \\
\cmidrule(lr){4-11} \cmidrule(lr){12-13}

& \textbf{Model} 
&  & \multicolumn{4}{c}{\textbf{ComCo}} & \multicolumn{4}{c}{\textbf{SimCo}} 
& \multirow{2}{*}{\textbf{VOC}}
& \multirow{2}{*}{\textbf{COCO}} \\

\cmidrule(lr){4-7} \cmidrule(lr){8-11}

&  &  & \textbf{2 obj.} & \textbf{3 obj.} & \textbf{4 obj.} & \textbf{5 obj.} 
& \textbf{2 obj.} & \textbf{3 obj.} & \textbf{4 obj.} & \textbf{5 obj.} 
&  &  \\
\midrule

\multirow{5}{*}{ViT-B/16}
& CLIP~\cite{radford2021learning}
&  & 77.55 & 80.31 & 81.41 & 80.22 
& 77.15 & 84.58 & 87.40 & 88.48 
& \multicolumn{1}{c}{78.56} & \multicolumn{1}{c}{53.94} \\

& MERU~\cite{desai2023hyperbolic}
&  & 72.90 & 77.25 & 78.15 & 77.34 
& 77.82 & 83.91 & 85.79 & 86.90 
& \multicolumn{1}{c}{79.50} & \multicolumn{1}{c}{54.39} \\

& ATMG$^\dagger$~\cite{ramasinghe2024accept}
&  & 45.91 & 45.97 & 45.80 & 45.82 
& 65.52 & 65.32 & 65.28 & 65.12 
& \multicolumn{1}{c}{72.22} & \multicolumn{1}{c}{46.81} \\

& HyCoCLIP~\cite{pal2024compositional}
&  & 72.90 & 73.22 & 73.51 & 72.90 
& 75.71 & 81.13 & 82.41 & 82.85 
& \multicolumn{1}{c}{80.43} & \multicolumn{1}{c}{58.12} \\

& \cellcolor{gray!10}\textbf{UNCHA (Ours)} 
& \cellcolor{gray!10} 
& \cellcolor{gray!10}\textbf{77.92} 
& \cellcolor{gray!10}\textbf{80.96} 
& \cellcolor{gray!10}\textbf{81.83} 
& \cellcolor{gray!10}\textbf{81.18} 
& \cellcolor{gray!10}\textbf{79.72} 
& \cellcolor{gray!10}\textbf{86.93} 
& \cellcolor{gray!10}\textbf{89.75} 
& \cellcolor{gray!10}\textbf{90.65} 
& \multicolumn{1}{c}{\cellcolor{gray!10}\textbf{82.14}} 
& \multicolumn{1}{c}{\cellcolor{gray!10}\textbf{59.43}} \\
\bottomrule
\end{tabular}

\label{tab:multiobj_rep_cls}
\vspace{-1em}
\end{table*}

\subsection{Training details}
To ensure a fair comparison, baseline models~\cite{pal2024compositional, radford2021learning, desai2023hyperbolic, ramasinghe2024accept} are reproduced under identical dataset and training configurations, while preserving the optimization settings specified in their original implementations. The batch size and total number of training iterations are fixed at 768 and 500,000, respectively. All models are trained on the Grounded Image-Text Pairs (GRIT)~\cite{peng2023kosmos} dataset, which contains 20.5 million grounded vision–language pairs and 35.9 million part-level annotations. Detailed descriptions of the settings and hyperparameters are provided in Sec.~S.1 of the supplementary material.

\subsection{Downstream tasks} 
\subsubsection{Zero-shot image classification} 
We conduct zero-shot classification experiments on 16 benchmark datasets as listed in Tab.~\ref{tab:zero_shot_cls}. We report Top-1 accuracy as the evaluation metric for all results following prior works~\cite{desai2023hyperbolic, radford2021learning}.
To evaluate scalability, we experiment with different sizes of vision encoders, ViT-S and ViT-B. For ATMG~\cite{ramasinghe2024accept}, we follow the original setup, computing similarity via averaged exterior angles instead of Lorentz or Euclidean inner products. This configuration is used for all downstream tasks. As shown in Tab.~\ref{tab:zero_shot_cls}, our method consistently outperforms prior approaches across all benchmark datasets, demonstrating generalization and robust performance on downstream tasks. 

\subsubsection{Zero-shot retrieval}  
For the retrieval task, we evaluate the model’s ability to retrieve the most relevant samples across modalities. Specifically, given an input image (or text), the model retrieves the Top-K text (or image) candidates from the collection, and the retrieval accuracy is computed accordingly. All experiments are conducted under the zero-shot setting using the COCO~\cite{lin2014microsoft} validation set and the Flickr30K~\cite{young2014image, karpathy2015deep} test set. As shown in Tab.~\ref{tab:retrieval_hier_imagenet}, our method shows steady performance, indicating its reliable cross-modal alignment capability across both benchmarks.

\subsubsection{Hierarchical Classification}  
To evaluate how well the model embeds hierarchical relationships in hyperbolic space, we adopt the hierarchy-aware metrics introduced in HyCoCLIP~\cite{pal2024compositional}. As shown in Tab.~\ref{tab:retrieval_hier_imagenet}, our model achieves consistently strong performance in hierarchical metrics, demonstrating its improved ability to preserve the structural hierarchy of the class labels within the embedding space, partly due to the uncertainty-guided alignment. More detailed explanations are in supplementary material Sec.~S.2.2.3.

\subsubsection{Zero-shot multi-label classification} 
We conduct multi-label classification experiments on the MS-COCO~\cite{lin2014microsoft} and VOC~\cite{everingham2010pascal} datasets, as shown in Tab.~\ref{tab:multiobj_rep_cls}. The evaluation metric is mean Average Precision (mAP). To further assess performance in more complex multi-object settings, we employed the ComCo and SimCo datasets~\cite{abbasi2025clip}. These datasets evaluate compositional understanding with images containing $N$ objects. ComCo features realistic object compositions, whereas SimCo provides synthetic scenes with diverse geometric shapes. For evaluation, we train a lightweight classifier on the embeddings and reported test-set classification mAP. As shown in Tab.~\ref{tab:multiobj_rep_cls}, UNCHA outperforms all baselines across both multi-label classification and multi-object representation benchmarks which indicate that our uncertainty-aware modeling provides a substantially stronger compositional understanding. These results highlight UNCHA’s ability to better disentangle object-level semantics and maintain robust alignment in  complex multi-object scenes.

\begin{figure}[!t]
    \centering
   \vspace{-0.7em}
    \includegraphics[width=0.4\textwidth]{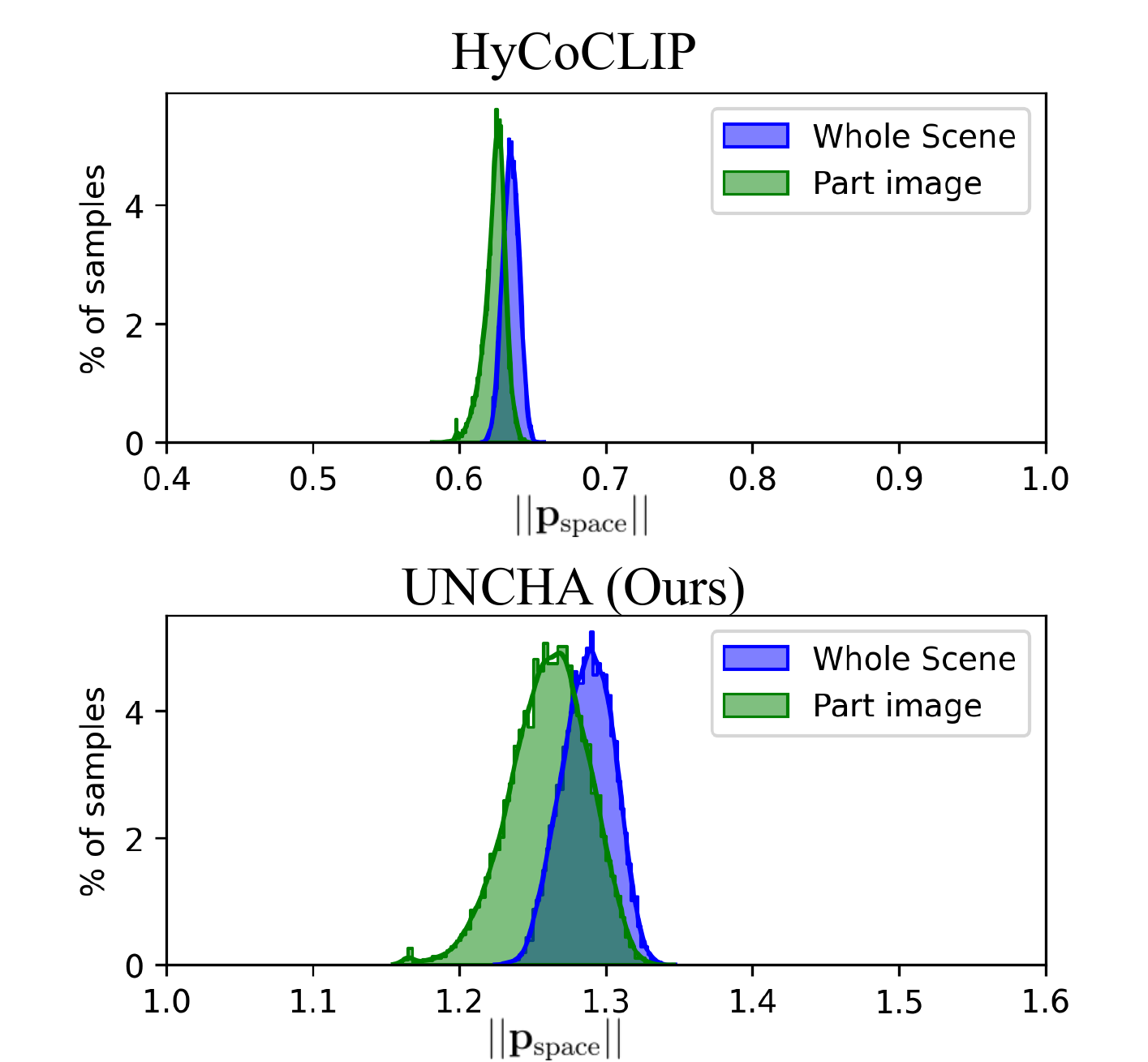} 
    \vspace{-0.5em}
    \caption{\textbf{Analysis of hyperbolic embedding.} Compared to HyCoCLIP~\cite{pal2024compositional}, whose hyperbolic embeddings exhibit a narrower range, UNCHA yields a more dispersed and structured distribution, reflecting richer use of the hyperbolic space.}
    \label{fig:analysis_norm}
   \vspace{-1.5em}
\end{figure}

\subsubsection{Part-level alignment with hard negatives} 
We evaluate part-level text–image matching using the benchmark derived from the densely annotated Densely Captioned Images~\cite{Urbanek_2024_CVPR}. The benchmark pairs cropped parts with their corresponding texts and introduces region-specific hard negatives to test fine-grained alignment. We report results on the `All Pick5' and `All Hard negs' in Tab.~\ref{tab:swap-hard}, which require the model not only to identify the correct pair among hard negative captions but also to produce a correct ordering between matching and non-matching pairs. UNCHA (Ours) achieves the highest performance among baselines, exhibiting substantial improvements in the `All Pick5' setting. This shows that our model effectively captures fine-grained part-whole distinctions, yielding better region-level visual-semantic alignment. 

\subsection{Analysis about hyperbolic space}  
We visualize the radii of hyperbolic embedding for $10,000$ ImageNet~\cite{russakovsky2015imagenet} images and their randomly cropped parts, shown in Fig.~\ref{fig:analysis_norm}. As noted in HyCoCLIP~\cite{pal2024compositional}, the embeddings of image and their parts often collapse into a narrowly concentrated region, yielding minimal separation between part and whole. In contrast, UNCHA produces a more distinctive and semantically structured geometry: part embeddings consistently lie closer to the origin than whole-scene embeddings, and the two distributions become clearly separated. This behavior results from the application of our uncertainty calibration and entropy regularizer. A more detailed analysis of hyperbolic space is provided in Sec.~S.2.5 of the supplementary material.

\subsection{Ablation study}
To assess the contribution of each component in our framework, we performed ablation experiments, each removing a distinct component. In Tab.~\ref{tab:ablation}, `w/o contrastive' removes the uncertainty-aware scaling from the global-local contrastive loss, while `w/o uncertainty' disables the uncertainty calibration in uncertainty-guided entailment loss. Finally, `w/o entropy' removes the entropy regularization from the uncertainty calibration module. The results demonstrate that all components of our method are essential. All experiments were conducted with ViT-S/16 architecture.

\section{Conclusion}
We propose UNCHA, a hyperbolic VLM that integrates part-to-whole representativeness, quantified as hyperbolic uncertainty, into both contrastive and entailment learning for hierarchy-aware compositional modeling. By further calibrating uncertainty using part-to-whole entailment relationships and an entropy based regularization term, our method enables efficient use of hyperbolic space and yields well-calibrated part-whole orderings. Extensive experiments on zero-shot classification, retrieval, and multi-label benchmarks, including complex multi-object scenes, demonstrate state-of-the-art performance, highlighting the importance of uncertainty-guided alignment for compositional understanding in vision-language learning.

\noindent \textbf{Acknowledgements}
This work was supported in part by Institute of Information \& communications Technology Planning \& Evaluation (IITP) grants funded by the Korea government(MSIT) [No.RS-2021-II211343, Artificial Intelligence Graduate School Program (Seoul National University) / No.RS-2025-02314125, Effective Human-Machine Teaming With Multimodal Hazy Oracle Models], the National Research Foundation of Korea(NRF) grants funded by the Korea government(MSIT) (Nos. RS-2022-NR067592, RS-2025-02263628), the AI Computing Infrastructure Enhancement (GPU Rental Support) User Support Program funded by the Ministry of Science and ICT (MSIT), Republic of Korea (No. RQT-25-120066), the BK21 FOUR program of the Education and Research Program for Future ICT Pioneers, Seoul National University and AI-Bio Research Grant through Seoul National University.

\clearpage
\appendix
\twocolumn[
\begin{center}
    { \Large \bfseries Supplementary Material for \\
\large Uncertainty-guided Compositional Alignment with Part-to-Whole Semantic Representativeness in Hyperbolic Vision-Language Models}
    \vspace{1em}
\end{center}
]

\renewcommand{\thefigure}{S.\arabic{figure}}
\renewcommand{\thetable}{S.\arabic{table}}
\renewcommand{\thesection}{S.\arabic{section}}
\renewcommand{\theequation}{S.\arabic{equation}}

\section{Implementation details}
\subsection{Model architecture} Our text encoder follows the CLIP~\cite{radford2021learning} design and uses a 12-layer, 512 dimensional Transformer~\cite{vaswani2017attention}. The maximum input length is set to $77$ tokens with a vocabulary size of $49,408$. For images, we adopt a Vision Transformer~\cite{dosovitskiy2020image} and experiment with two capacity configurations, ViT-S and ViT-B~\cite{touvron2021training, chen2020towards}, both using a patch size of 16. These architectural choices are consistent with prior works~\cite{desai2023hyperbolic, pal2024compositional}. During training, we apply the same image augmentations as OpenCLIP~\cite{ilharco_gabriel_2021_5143773}, including random cropping, random grayscale conversion, and random color jittering, and resize all images to \(224 \times 224\).

\subsection{Model initialization} The curvature of Lorentz space is initialized to $\kappa=1.0$ and treated as a learnable parameter, while being clamped in $[0.1, 10.0]$ for numerical stability. The final learned value converges to $\kappa=0.1$ , consistent with those used in prior hyperbolic methods~\cite{desai2023hyperbolic, pal2024compositional, ramasinghe2024accept}. Before projecting representations onto the Lorentz model, we apply learnable scaling factors to image and text vectors. These scalars are initialized as $c_{\text{img}} = c_{\text{txt}} = \frac{1}{\sqrt{512}}$, following prior work~\cite{desai2023hyperbolic, pal2024compositional}. The temperature parameters are also learnable. The global-local logit scale $\tau_{gl}$ is initialized to $0.06$, while the local and global logit scales, $\tau_{l}$ and $\tau_{g}$, are initialized to $0.07$. All temperature values are clipped at a minimum of $0.01$. Values of $\eta$ parameter are set to $\eta_{intra} = 1.2$ for intra-modality entailments and $\eta_{inter} = 0.7$ for inter-modal entailments (Eq.14) and $K=0.1$ (Eq.13), following~\cite{pal2024compositional}. In Eq.14, we set $\alpha=0.1$. For Eq.17, the weighting coefficients are $\lambda_{1} =0.5$ and $\lambda_{2} =10.0$. In Eq.18, we use $\lambda_{ent} =0.2$.  

\subsection{Optimizer and hardware}

Our model is trained for 500K steps using four A100 GPUs with a batch size of 768. We employ the AdamW optimizer~\cite{loshchilov2017decoupled}, setting $\beta_1=0.9, \beta_2=0.98$, and a weight decay of $0.2$. The decay is excluded for the learnable parameters, including the temperature parameters, curvature, and the scaling factors $c_{\text{img}}$ and $ c_{\text{txt}}$. We adopt a cosine learning-rate scheduler~\cite{loshchilov2016sgdr} with a maximum learning rate of $5 \times 10^{-4}$, with a 4k-step linear warm-up period.

\section{Additional details on experiments}
\subsection{Training details on other models} We employ  CLIP~\cite{radford2021learning}, MERU~\cite{desai2023hyperbolic}, and HyCoCLIP~\cite{pal2024compositional} models trained on the Grounded Image-Text Pairs (GRIT) dataset~\cite{peng2023kosmos}, using the reproduced version released by~\cite{pal2024compositional}. For CLIP~\cite{radford2021learning} and MERU~\cite{desai2023hyperbolic}, we adopt the variants trained without part images, as their original training pipelines do not incorporate part-level data and prior work~\cite{pal2024compositional} reports that including part images does not lead to performance improvements. The GRIT dataset contains $20.5$ million grounded vision-language pairs and $35.9$ million box-level annotations describing objects within each scene, derived from the larger COYO-700M corpus~\cite{kakaobrain2022coyo700m}. In addition, we train ATMG~\cite{ramasinghe2024accept} on the same GRIT dataset using a batch size of $768$ for 500K iterations, preserving the optimization settings specified in their original implementation. 

\subsection{Downstream tasks} 
\subsubsection{Zero-shot image classification} For MERU~\cite{desai2023hyperbolic}, HyCoCLIP~\cite{pal2024compositional} and UNCHA (Ours), similarity between text and image embedding is computed with Lorentzian inner product. For CLIP~\cite{radford2021learning}, similarity is measured using the Euclidean inner product, while for ATMG~\cite{ramasinghe2024accept}, we adopt its exterior angle-based similarity. The same similarity formulation for each model is consistently applied across all remaining downstream tasks. In zero-shot image classification, we treat the label set as a collection of text queries~\cite{elhoseiny2013write} and apply prompt ensembling for each label by encoding multiple prompt variants and averaging their embeddings before generating the final textual representations, following previous works~\cite{desai2023hyperbolic, pal2024compositional, ramasinghe2024accept}. Using these embedded text queries, we compute image-text similarities and report top-1 accuracy averaged over classes. 

\subsubsection{Zero-shot retrieval} In zero-shot text-to-image retrieval, we compare every text caption embedding against all image embeddings and sort the images in descending order of similarity. The same procedure is applied symmetrically for image-to-text retrieval. We compute recall@K for both directions using the ground-truth associations provided by COCO~\cite{lin2014microsoft} and Flickr30K~\cite{young2014image, karpathy2015deep}, where a retrieval is counted as correct if at least one paired item appears within the top-K results. All recall metrics are averaged over the full set of queries to produce final results. 

\subsubsection{Hierarchical classification} For hierarchical classification task, we follow the prior work~\cite{pal2024compositional} and use the WordNet hierarchy~\cite{miller1995wordnet} of the ImageNet class labels~\cite{deng2009imagenet, russakovsky2015imagenet}. The Tree-Induced Error (TIE) quantifies how far the predicted label is from the ground-truth label within the given tree. The Lowest Common Ancestor (LCA) error captures how far each label is from their deepest shared ancestor, defined as the sum of the edge-weighted distances from the predicted and true labels to the LCA. Set-based metrics compare the ancestor sets of the predicted and true labels: using all ancestor nodes for each label, we compute Jaccard similarity, hierarchical precision, and hierarchical recall based on their set intersection.

\subsubsection{Zero-shot multi-label classification} 
\label{para:multi-label-re}
\paragraph{Multi-label classification.} We perform multi-label classification experiments on the MS-COCO~\cite{lin2014microsoft} and VOC~\cite{everingham2010pascal} datasets and report performance using mean Average Precision (mAP). This task evaluates whether the VLM can correctly predict the set of classes present in each image by comparing its predictions against the binary ground-truth labels. Because the baseline models include both hyperbolic and Euclidean variants, their similarity score ranges differ substantially: Euclidean VLMs typically output similarities within $[0, 1]$, whereas hyperbolic similarity scores generally fall at or below –10. To ensure a fair comparison across models, we apply an additional normalization step to the similarity scores before computing the evaluation metrics.

\paragraph{Multi-object representation.} This benchmark is designed to evaluate more complex multi-object scenarios using the ComCo and SimCO datasets~\cite{abbasi2025clip}. As described in~\cite{abbasi2025clip}, this setting allows us to assess how well a VLM’s image encoder represents individual objects within multi-object scenes and to analyze whether its representations exhibit biases with respect to object size. ComCo consists of images containing realistic 3D asset objects, such as cars or airplanes, arranged in sets of $N$, while SimCo contains synthetic 3D assets such as blue spheres, cones, and other primitive shapes. In both datasets, each image contains between two and five objects, so the labels `2 obj.', `3 obj.' in Tab.5 of the main text refers to sets of images that contain exactly two or three objects, respectively. These images include various combinations of object sizes and spatial arrangement. For instance, `3 obj.' set contain one large object and two smaller objects in different location. A separate classifier is trained for each set on top of the features produced by the VLM's image encoder, grouped by the number of objects. The model is evaluated on its ability to distinguish all components across different sizes and positions, and at test time we assess whether the trained classifier can correctly identify each component in response to new text queries. Extended results evaluated with the ViT-S backbone are provided in Tab.~\ref{tab:multi-obj-full}.

\subsubsection{Part-level alignment with hard negatives} 
\label{sec:part-level-alignment}
This benchmark, introduced in~\cite{Urbanek_2024_CVPR}, evaluates whether a VLM can correctly associate captions with the appropriate image subregions when multiple submasks and captions exist for the same image, using the 7,805 images from the summarized Densely Captioned Images (sDCI) dataset. The original DCI dataset provides dense textual annotations, including multiple captions, subcaptions, and visual descriptions per image. To align these annotations with CLIP-style input constraints, all LLM-generated captions are truncated to 77 tokens to form sDCI. Each image contains several subcrops, each paired with one or more summarized captions as well as LLM-generated negatives. Retrieval-style evaluations are constructed by placing multiple subcrops and captions from the same image within a single batch, requiring the model to identify which caption corresponds to which region. 

We report the result of `All Pick5-SCM', `All Pick5-Neg', and `All-Hard Negs' in the main paper, and include all metrics below, tested with both ViT-B, ViT-S at Tab.~\ref{tab:part-level}. In `All-SCM', one summarized caption is paired with each subcrop, and the model must identify the caption that describes that specific region, distinguishing it from captions corresponding to other subcrops of the same image as well as from other in-batch captions. In `All-Neg', each subcrop’s caption is evaluated against an LLM-generated negative to test positive-negative discrimination. The `All Pick5-SCM' setting follows the structure of `All-SCM' but uses five captions per subcrop, with success only if the correct caption scores higher than all positives from other images. In `All Pick5-Neg', five summarized captions are paired with each subcrop, and the model succeeds only if all positives score above the negative. In `Base-Neg', only full images (no subcrops) are used, and each image is paired with its LLM-generated negative caption to test the models' ability to distinguish between an LLM generated caption and its corresponding LLM-generated negative. Finally, `All-Hard Negs' follows the same setup as `All-Neg' but replaces the negative caption with the hardest (highest-scoring) LLM-generated negative across the entire negative pool.

\subsection{Additional ablation study}

\subsubsection{Ablation study on hyperbolic radius}
As discussed in the main paper, for a point $\mathbf{x} \in \mathbb{L}^n$, we define the
uncertainty $u$ using the Euclidean $\ell_2$ norm of $\mathbf{x}$, since this norm is
monotonically proportional to its hyperbolic radius. We represent a point
$\mathbf{x} \in \mathbb{R}^{n+1}$ in the Lorentz model using its time--space decomposition:
\begin{equation}
\mathbf{x} = [x_{\text{time}}, \mathbf{x}_{\text{space}}], 
\qquad 
x_{\text{time}} \in \mathbb{R}, \; \mathbf{x}_{\text{space}} \in \mathbb{R}^n
\end{equation}
The origin of the hyperboloid corresponds to the point
$\mathbf{o} = [\sqrt{1/\kappa}, \mathbf{0}]$. Therefore, the hyperbolic radius, defined as the geodesic distance between
$\mathbf{x}$ and the origin, can be calculated as:
\begin{equation}
\begin{aligned}
d_{\mathbb{L}}(\mathbf{x}, \mathbf{o})
&= \sqrt{\frac{1}{\kappa}} \,
   \cosh^{-1}\!\left( -\kappa \langle \mathbf{x}, \mathbf{o} \rangle_{\mathbb{L}} \right) \\
&= \sqrt{\frac{1}{\kappa}} \,
   \cosh^{-1}\!\left( x_{\text{time}} \sqrt{\kappa} \right)
\end{aligned}
\end{equation}
where we used the Lorentzian inner product
\begin{equation}
\langle \mathbf{x}, \mathbf{o} \rangle_{\mathbb{L}}
= -x_{\text{time}} \sqrt{\frac{1}{\kappa}}
\end{equation}
To obtain an explicit expression, we use the hyperboloid constraint:
\begin{equation}
\langle \mathbf{x}, \mathbf{x} \rangle_{\mathbb{L}}
= -x_{\text{time}}^2 + \|\mathbf{x}_{\text{space}}\|^2_2
= -\frac{1}{\kappa},
\end{equation}
which implies
\begin{equation}
x_{\text{time}}
= \sqrt{\|\mathbf{x}_{\text{space}}\|^2_2 + \frac{1}{\kappa}}
\end{equation}
As we mentioned in preliminaries of main text, we only parameterize the space component of $\mathbf{x}$. Hence, the Euclidean norm satisfies $\lVert \mathbf{x}_{\text{space}}\rVert_2 \equiv \lVert \mathbf{x}\rVert_2$ for our parameterization.  Therefore, the geodesic distance (hyperbolic radius) from the origin to a point $\mathbf{x} \in \mathbb{R}^D$ is given by:
\begin{equation}
\begin{aligned}
d_{\mathbb{L}}(\mathbf{x}, \mathbf{o})
= \frac{1}{\sqrt{\kappa}} 
  \cosh^{-1}\!\left(
    \sqrt{1 + \kappa \lVert \mathbf{x} \rVert_2^2}
  \right)
\end{aligned}
\label{eq:hyperbolic-radius}
\end{equation}
This expression reveals that the hyperbolic radius is closely related to the Euclidean norm of $\mathbf{x}$,  $\lVert \mathbf{x} \rVert_2$. 

For small  $\mathbf{x}$,  $\lVert \mathbf{x} \rVert_2$, we have the approximation
\begin{equation}
    \sqrt{1 + \kappa \lVert \mathbf{x} \rVert_2^2}
    \approx 
    1 + \frac{\kappa}{2}\lVert \mathbf{x} \rVert_2^2
\end{equation}
and using $\cosh^{-1}(1+u) \approx \sqrt{2u}$, it follows that
\begin{equation}
     d_{\mathbb{L}}(\mathbf{x}, \mathbf{o}) \approx \lVert \mathbf{x} \rVert_2
\end{equation}
showing that the hyperbolic radius grows approximately proportionally to the Euclidean norm for small $\lVert \mathbf{x} \rVert_2$.

For large norms, using $\cosh^{-1}(t) \approx \log(2t)$, the radius behaves as: 
\begin{equation}
     d_{\mathbb{L}}(\mathbf{x}, \mathbf{o}) 
    \approx \frac{1}{\sqrt{\kappa}}
    \log\!\left( 2 \sqrt{\kappa} \lVert \mathbf{x} \rVert_2 \right)
\end{equation}
indicating a transition to logarithmic growth. Overall, the hyperbolic radius is approximately proportional to the Euclidean norm for small $\lVert \mathbf{x} \rVert_2$, but grows logarithmically for large $\lVert \mathbf{x} \rVert_2$. This monotonic relationship validates the use of the Euclidean norm of $\mathbf{x}$ as a proxy for its hyperbolic radius. This enables us to avoid the unnecessary hyperbolic computations while preserving the same ordering. The ablation result obtained when training directly with the hyperbolic radius in Eq.~\ref{eq:hyperbolic-radius} is reported in Tab.~\ref{tab:ablation_hyperbolic_radius}, showing slightly reduced performance compared to our full model. This confirms that our Euclidean norm proxy provides an effective surrogate for the hyperbolic radius, enabling more reliable uncertainty estimation during training.  

\begin{table}[t]
\centering
\small
\setlength{\tabcolsep}{5pt}
\renewcommand{\arraystretch}{0.95}

\caption{\textbf{Ablation study on hyperbolic radius.} Replacing our Euclidean-norm surrogate with the explicit hyperbolic radius slightly degrades both classification and retrieval performance. Bold numbers indicate the best within each task group.}

\begin{tabular}{lccccc}
\toprule
& \multicolumn{3}{c}{\cellcolor{lightblue}\textbf{Classification}}
& \multicolumn{2}{c}{\cellcolor{lightcyan}\textbf{Retrieval}} \\
\cmidrule(lr){2-4} \cmidrule(lr){5-6}

\textbf{Model}
& \rotatebox{90}{Gen.}
& \rotatebox{90}{Fine.}
& \rotatebox{90}{MISC.}
& \rotatebox{90}{Text}
& \rotatebox{90}{Image} \\
\midrule

\cellcolor{gray!10}\textbf{Ours (full)}
& \cellcolor{gray!10}\textbf{68.98}
& \cellcolor{gray!10}\textbf{25.53}
& \cellcolor{gray!10}\textbf{27.55}
& \cellcolor{gray!10}\textbf{83.80}
& \cellcolor{gray!10}\textbf{73.90} \\

with $ d_{\mathbb{L}}(\mathbf{x}, \mathbf{o})$ 
& 67.41
& 24.81
& 25.55
& 79.43
& 72.00 \\
\bottomrule
\end{tabular}

\label{tab:ablation_hyperbolic_radius}
\end{table}

\subsubsection{Analysis experiments}
\paragraph{Analysis of uncertainty modeling.}
In Fig. 4(a), we investigate how uncertainty reflects the semantic representativeness of local regions within an image. To this end, we randomly crop multiple patches from a single image and compute the uncertainty for each patch. The patches are then arranged according to their uncertainty values, from low to high, progressing from the top-left to the bottom-right. We observe that patches with low uncertainty tend to correspond to semantically meaningful and well-aligned regions, such as prominent objects or structurally informative parts of the scene. In contrast, patches with high uncertainty are often blurred, textureless, or less informative, making them less representative of the overall scene. This qualitative observation suggests that our uncertainty measure effectively captures how well a local region aligns with the global semantics of the image. Additional results on uncertainty-based ordering are provided in Fig.~\ref{fig:norm-sorting-1}.
In Fig. 4(b), we further provide a quantitative analysis of this behavior using a subset of ImageNet~\cite{russakovsky2015imagenet}. For each image, we compute the semantic similarity between each cropped part and the corresponding whole image, and examine its relationship with the estimated uncertainty. The resulting scatter plot reveals a strong negative correlation (Corr = -0.739), indicating that parts with higher semantic similarity to the whole tend to have lower uncertainty, while less representative parts exhibit higher uncertainty. This consistent trend supports the interpretation that our uncertainty measure serves as a reliable proxy for semantic representativeness, which is crucial for accurate and robust part-level alignment.

\subsection{Additional experimental results}
\subsubsection{Part-level alignment with hard negatives}
\paragraph{Experimental setting.} The experimental setting is described in Sec.4.2.5 of the main text and further detailed in Sec.~\ref{sec:part-level-alignment}.
\paragraph{Experimental results.} Tab.~\ref{tab:part-level} presents the results for the part-level alignment benchmark with hard negatives across evaluation settings described in Sec.~\ref{sec:part-level-alignment}. Across both ViT-S/16 and ViT-B/16 backbones, UNCHA (Ours) consistently achieves the best or second-best performance in nearly every setting. The gains are especially noticeable in the more challenging `All Pick5-SCM' and `All Pick5-Neg' settings, where multiple positives per sub-crop make the matching task substantially harder. Even in the `All-Hard Negs' setting, where each sub-crop must be distinguished from the hardest negative caption selected from the entire LLM-generated negative pool, UNCHA achieves the best performance, demonstrating its robustness against challenging negative distractors. This result indicates that UNCHA (Ours) effectively identifies and differentiates distinct subregions within an image, demonstrating its ability to understand images in a more fine-grained manner.

\begin{table*}[t]
\centering
\small
\setlength{\tabcolsep}{7pt}

\caption{\textbf{Full results of part-level alignment with hard negatives.} 
Comparison across all settings of part-level alignment with hard negatives for ViT-S and ViT-B. UNCHA (Ours) consistently outperforms prior models, including the challenging `All Pick5' and `All-Hard Negs' settings, demonstrating its strong capability in accurately identifying and distinguishing fine-grained visual regions within images.}

\begin{tabular}{llcccccc}
\toprule
& & \multicolumn{2}{c}{\cellcolor{lightblue}\textbf{All}} 
& \multicolumn{2}{c}{\cellcolor{lightcyan}\textbf{All Pick5}} 
& \multicolumn{1}{c}{\cellcolor{lightgreen}\textbf{Base}} 
& \multicolumn{1}{c}{\cellcolor{lightmint}\textbf{All}} \\
\cmidrule(lr){3-4} \cmidrule(lr){5-6} \cmidrule(lr){7-8}

& \textbf{Model}
& \textbf{SCM} & \textbf{Neg} 
& \textbf{SCM} & \textbf{Neg} 
& \textbf{Neg} & \textbf{Hard Negs} \\
\midrule

\multirow{5}{*}{\textbf{ViT-S/16}}
& CLIP~\cite{radford2021learning} & 39.87 & {63.60} & \underline{12.52} & \underline{23.88} & \underline{82.41} & \underline{57.31} \\
& ATMG~\cite{ramasinghe2024accept} & {40.45} & 61.51 & 12.30 & 22.29 & 73.15 & 55.79 \\
& MERU~\cite{desai2023hyperbolic} & \underline{40.81}& \textbf{64.18} & 12.23 & {23.81} & 79.63 & {56.30} \\
& HyCoCLIP~\cite{pal2024compositional} & 36.61 & 60.13 & 10.85 & 22.29 & 80.56 & 52.03 \\
& \cellcolor{gray!10}\textbf{UNCHA (Ours)} 
& \cellcolor{gray!10}\textbf{41.10} 
& \cellcolor{gray!10}\underline{63.89}
& \cellcolor{gray!10}\textbf{12.88} 
& \cellcolor{gray!10}\textbf{25.04}
& \cellcolor{gray!10}\textbf{83.33}
& \cellcolor{gray!10}\textbf{57.45} \\
\midrule

\multirow{5}{*}{\textbf{ViT-B/16}}
& CLIP~\cite{radford2021learning} & 39.22 & 59.33 & \underline{13.10} & {22.94} & 74.07 & {52.89} \\
& ATMG~\cite{ramasinghe2024accept} & \textbf{40.38} & {62.08} & 12.23 & 23.08 & \textbf{82.41} & 53.91 \\
& MERU~\cite{desai2023hyperbolic}& \underline{40.09} & \textbf{62.37} & 12.59 & 20.69 & \underline{81.48} & \underline{54.56} \\
& HyCoCLIP~\cite{pal2024compositional} & 35.96 & 60.78 & 11.65 & \underline{23.52} & 75.93 & 53.33 \\
& \cellcolor{gray!10}\textbf{UNCHA (Ours)}
& \cellcolor{gray!10}39.58 
& \cellcolor{gray!10}\underline{62.23}
& \cellcolor{gray!10}\textbf{13.53} 
& \cellcolor{gray!10}\textbf{23.81}
& \cellcolor{gray!10}{80.56}
& \cellcolor{gray!10}\textbf{56.51} \\
\bottomrule
\end{tabular}
\label{tab:part-level}
\end{table*}

\subsubsection{Multi-object representation}
\paragraph{Experimental setting.} The experimental setting is described in Sec.4.2.4 of the main text and further detailed in Sec.~\ref{para:multi-label-re}.
\paragraph{Experimental results.} We extend the multi-object representation experiments from the main paper by additionally evaluating ViT-S models. As presented in Tab.~\ref{tab:multi-obj-full}, UNCHA (Ours) consistently achieves superior performance across diverse object counts and datasets. This reflects its ability to reliably represent and distinguish individual objects within complex multi-object scenes, demonstrating strong fine-grained and compositional understanding. 

\begin{table*}[t]
\centering
\small
\setlength{\tabcolsep}{6pt}

\caption{\textbf{Multi-object representation performance on ComCo and SimCo (mAP).} UNCHA (Ours) generally outperforms all baselines across object counts and datasets in the extended ViT-S and ViT-B evaluation (Tab.~\ref{tab:multi-obj-full}), demonstrating strong fine-grained and compositional understanding in complex multi-object scenes.} 
\vspace{0.7em}

\begin{tabular}{llcccccccc}
\toprule
\multicolumn{2}{c}{} 
& \multicolumn{4}{c}{\cellcolor{lightblue}\textbf{ComCo}} 
& \multicolumn{4}{c}{\cellcolor{lightgreen}\textbf{SimCo}} \\
\cmidrule(lr){3-6} \cmidrule(lr){7-10}

& & \textbf{2 obj} & \textbf{3 obj} & \textbf{4 obj} & \textbf{5 obj}
& \textbf{2 obj} & \textbf{3 obj} & \textbf{4 obj} & \textbf{5 obj} \\
\midrule

\multirow{5}{*}{\textbf{ViT-S/16}}
& CLIP~\cite{radford2021learning}
& \textbf{69.59} & \textbf{71.97} & \underline{72.44} & \underline{72.06}
& 72.49 & \underline{80.05} & \underline{82.45} & \underline{82.65} \\

& MERU~\cite{desai2023hyperbolic}
& 67.42 & 69.31 & 70.04 & 69.60
& 71.69 & 78.56 & 80.65 & 81.20 \\

& ATMG~\cite{ramasinghe2024accept}
& 44.01 & 43.94 & 44.12 & 43.97
& 62.17 & 63.02 & 61.83 & 62.00 \\

& HyCoCLIP~\cite{pal2024compositional}
& 64.47 & 65.67 & 66.37 & 65.74
& \underline{72.91} & 78.25 & 79.55 & 79.43 \\

& \cellcolor{gray!10}\textbf{UNCHA (Ours)}
& \cellcolor{gray!10}\underline{68.91}
& \cellcolor{gray!10}\underline{71.54}
& \cellcolor{gray!10}\textbf{72.90}
& \cellcolor{gray!10}\textbf{72.58}
& \cellcolor{gray!10}\textbf{74.41}
& \cellcolor{gray!10}\textbf{81.79}
& \cellcolor{gray!10}\textbf{83.55}
& \cellcolor{gray!10}\textbf{83.13} \\
\midrule

\multirow{5}{*}{\textbf{ViT-B/16}}
& CLIP~\cite{radford2021learning}
& \underline{77.55} & \underline{80.31} & \underline{81.41} & \underline{80.22}
& 77.15 & \underline{84.58} & \underline{87.40} & \underline{88.48} \\

& MERU~\cite{desai2023hyperbolic}
& 72.90 & 77.25 & 78.15 & 77.34
& \underline{77.82} & 83.91 & 85.79 & 86.90 \\

& ATMG~\cite{ramasinghe2024accept}
& 45.91 & 45.97 & 45.80 & 45.82
& 65.52 & 65.32 & 65.28 & 65.12 \\

& HyCoCLIP~\cite{pal2024compositional}
& 72.90 & 73.22 & 73.51 & 72.90
& 75.71 & 81.13 & 82.41 & 82.85 \\

& \cellcolor{gray!10}\textbf{UNCHA (Ours)}
& \cellcolor{gray!10}\textbf{77.92}
& \cellcolor{gray!10}\textbf{80.96}
& \cellcolor{gray!10}\textbf{81.83}
& \cellcolor{gray!10}\textbf{81.18}
& \cellcolor{gray!10}\textbf{79.72}
& \cellcolor{gray!10}\textbf{86.93}
& \cellcolor{gray!10}\textbf{89.75}
& \cellcolor{gray!10}\textbf{90.65} \\

\bottomrule
\label{tab:multi-obj-full}
\end{tabular}
\end{table*}

\subsubsection{Zero-shot semantic segmentation}
\label{sec:semantic-seg}
\paragraph{Experimental setting.} 
Zero-shot semantic segmentation refers to benchmark settings where additional attention-modulation methods (such as SCLIP~\cite{wang2024sclip} and NACLIP~\cite{hajimiri2025pay}) are integrated into the model to extract not only class-level features but also the dense features produced by the backbone. Using these dense features, the model performs classification by comparing them against the class texts from existing datasets. In our experiments, we first use NACLIP to extract dense features and then compute their similarity to class texts, evaluating how accurately the model localizes fine-grained regions based on the  mIoU metric. However, semantic segmentation is substantially more challenging than standard image classification, so instead of relying solely on text–image matching as in typical classification, we further reduce the modality mismatch by extrapolating the text embeddings from the root of the hyperbolic space for all hyperbolic-based models.
\paragraph{Experimental results.}
As shown in Tab.~\ref{tab:zero-seg} and Fig.~\ref{fig:voc_vis_1}–\ref{fig:voc_vis_2}, our method consistently achieves strong performance across both the ViT-S and ViT-B backbones, indicating that it captures fine-grained details in images more effectively than existing approaches.
Furthermore, the results demonstrate that our method produces more coherent region assignments and reliably handles scenes containing multiple objects, correctly separating and allocating each instance. Taken together, these observations highlight the robustness and strong fine-grained awareness capability of our model in zero-shot segmentation settings.

\begin{table}[t]
\centering
\small
\setlength{\tabcolsep}{8pt}
\vspace{0em}
\caption{\textbf{Zero-shot segmentation performance on VOC21.} UNCHA (Ours) model achieves the highest mIoU on both the ViT-S/16 and ViT-B/16 backbones, showing clear improvements over prior methods. This result demonstrates that our hyperbolic alignment enables the model to effectively capture fine-grained region-level features.}
\vspace{0.3em}
\label{tab:zero-seg}
\begin{tabular}{lcc}
\toprule
\multicolumn{3}{c}{\cellcolor{lightblue}\textbf{VOC 21 dataset}} \\
\midrule
\textbf{Model} & \textbf{ViT-S/16} & \textbf{ViT-B/16} \\
\midrule
CLIP       & 36.02 & \underline{28.47} \\
MERU       & 36.18 & 26.05 \\
AtMG       &  7.63 &  6.51 \\
HyCoCLIP   & \underline{36.79} & 26.03 \\
\cellcolor{gray!10}\textbf{UNCHA (Ours)}
& \cellcolor{gray!10}\textbf{39.03}
& \cellcolor{gray!10}\textbf{32.28} \\
\bottomrule
\end{tabular}

\end{table}

\subsubsection{Bounding box classification}
\paragraph{Experimental setting.} Bounding box classification evaluates a model’s ability to recognize objects within localized regions using only textual descriptions. Following prior work~\cite{xie2025fg, jing2024fineclip}, we crop bounding boxes from COCO-val2017~\cite{lin2014microsoft}, LVIS~\cite{gupta2019lvis}, and Open Images~\cite{kuznetsova2020open} and classify them in a zero-shot manner.
\paragraph{Experimental results.} We report Top-1 and Top-5 accuracy in Tab.~\ref{tab:box-level zeroshot}. These results demonstrate that UNCHA (Ours) achieves consistently superior performance across all datasets, COCO, LVIS, and OpenImages, showing large gains over existing approaches. The improvements are particularly prominent in the Top-1 accuracy, reaching margins as high as 32.89\%, which highlights the model's ability to precisely associate localized visual regions with their corresponding textual concepts under zero-shot settings. This suggests that UNCHA (Ours) produces representations that remain stable and discriminative even when object regions are tightly cropped, where contextual cues are minimized. 
\begin{table*}[t]
\vspace{-0.5cm}
\centering
\small
\setlength{\tabcolsep}{6pt}

\caption{\textbf{Box-level zero-shot classification accuracy on COCO~\cite{lin2014microsoft}, LVIS~\cite{gupta2019lvis}, and OpenImages~\cite{kuznetsova2020open}.} We report Top-1 and Top-5 accuracy. UNCHA (Ours) achieves consistently superior performance across all datasets, showing substantial improvements over CLIP~\cite{radford2021learning}, MERU~\cite{desai2023hyperbolic}, ATMG~\cite{ramasinghe2024accept}, and HyCoCLIP~\cite{pal2024compositional} with Top-1 gains reaching up to $32.89\%$. These results indicate that our hyperbolic alignment mechanism enables more reliable region-level grounding and captures part-whole semantic structure more faithfully than prior baselines.}

\begin{tabular}{llccccccccc}
\toprule
& & \multicolumn{2}{c}{\cellcolor{lightblue}\textbf{COCO}}
& \multicolumn{2}{c}{\cellcolor{lightcyan}\textbf{LVIS }}
& \multicolumn{2}{c}{\cellcolor{lightgreen}\textbf{OpenImages}} \\
\cmidrule(lr){3-4} \cmidrule(lr){5-6} \cmidrule(lr){7-8}

& \textbf{Model}
& \textbf{Top-1} & \textbf{Top-5}
& \textbf{Top-1} & \textbf{Top-5}
& \textbf{Top-1} & \textbf{Top-5} \\
\midrule
\multirow{5}{*}{\textbf{ViT-S/16}}
& CLIP      & 34.98 & 60.74 &  5.81 & 13.97 & 13.81 & 35.76 \\
& MERU      & {43.51} & {66.77} &  6.43 & 15.06 & 16.51 & 41.26 \\
& ATMG      & 19.24 & 34.85 &  5.45 & 13.49 &  9.72 & 26.28 \\
& HyCoCLIP  & \underline{45.36} & \underline{73.17} & \underline{11.12} & \underline{25.28} & \underline{20.79} & \underline{47.57} \\
& \cellcolor{gray!10}\textbf{Ours}
& \cellcolor{gray!10}\textbf{51.57}
& \cellcolor{gray!10}\textbf{77.11}
& \cellcolor{gray!10}\textbf{13.65}
& \cellcolor{gray!10}\textbf{29.03}
& \cellcolor{gray!10}\textbf{24.36}
& \cellcolor{gray!10}\textbf{53.26} \\
\midrule
\multirow{5}{*}{\textbf{ViT-B/16}}
& CLIP      & 35.22 & 62.84 &  6.84 & 16.16 & 14.90 & 38.18 \\
& MERU      & 44.55 & 68.10 &  7.41 & 16.37 & 18.14 & 42.23 \\
& ATMG      & 21.19 & 37.61 &  6.19 & 14.84 & 10.52 & 29.09 \\
& HyCoCLIP  & \underline{47.88} & \underline{74.79} & \underline{12.92} & \underline{27.31} & \underline{22.16} & \underline{48.78} \\
& \cellcolor{gray!10}\textbf{Ours}
& \cellcolor{gray!10}\textbf{54.14}
& \cellcolor{gray!10}\textbf{79.03}
& \cellcolor{gray!10}\textbf{17.17}
& \cellcolor{gray!10}\textbf{33.21}
& \cellcolor{gray!10}\textbf{23.81}
& \cellcolor{gray!10}\textbf{52.53} \\

\bottomrule
\label{tab:box-level zeroshot}
\end{tabular}
\vspace{-0.5cm}
\end{table*}


\subsection{Analysis}
\subsubsection{Hyperbolic embedding analysis}
\begin{figure}[!t]
    \centering
\vspace{-0.3cm}
     \includegraphics[width=1\linewidth]{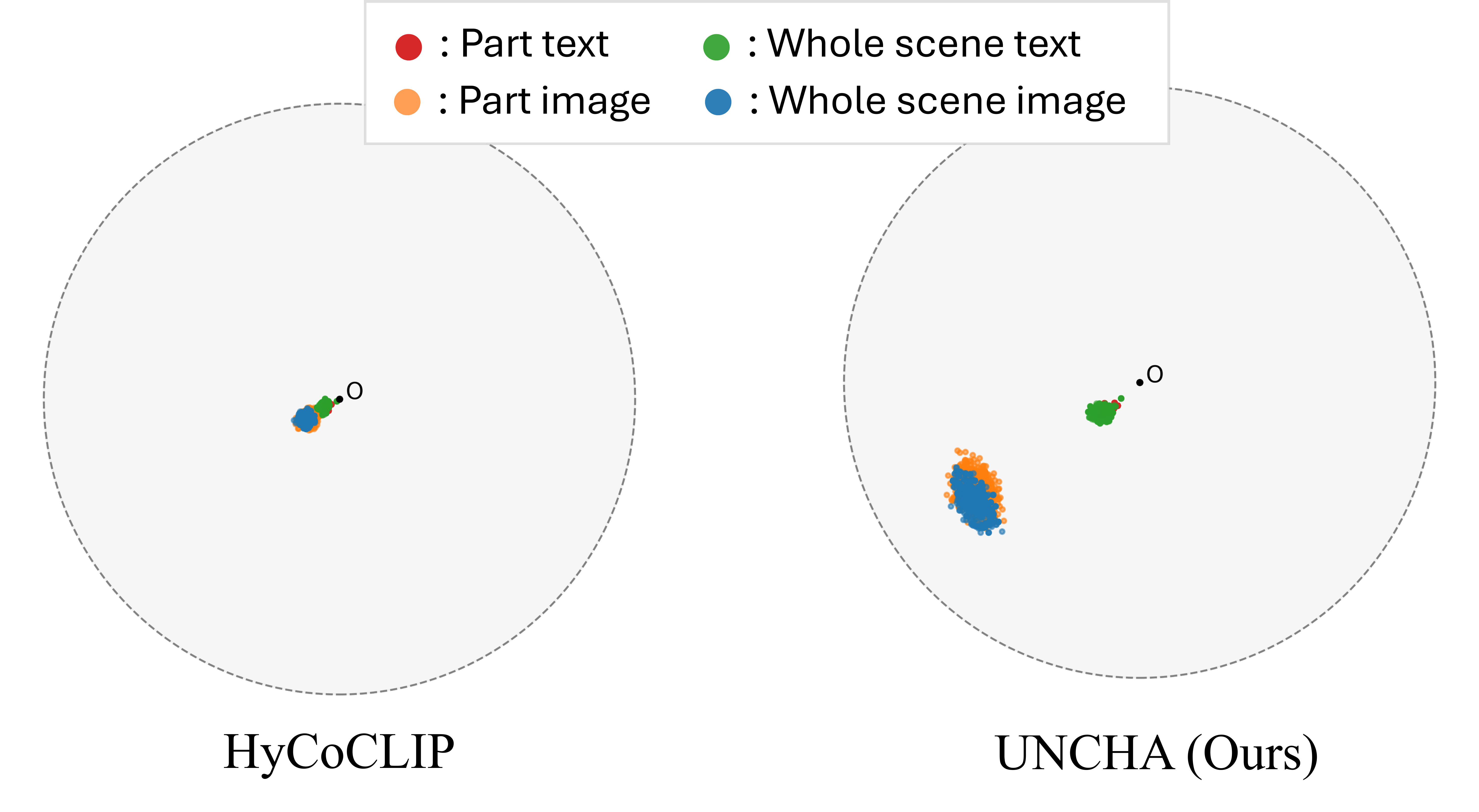}
     \vspace{-0.2cm}
    \caption{\textbf{Hyperbolic embedding visualization using HoroPCA.}
    On the COCO dataset~\cite{lin2014microsoft}, we compare the hyperbolic embeddings of our model with those of HyCoCLIP~\cite{pal2024compositional}. While HyCoCLIP embeddings are largely concentrated near the origin, ours are distributed farther away, enabling a broader and more effective utilization of the hyperbolic space.}
    \label{fig:coco_distrib}
    \setlength{\intextsep}{5pt} 
    \vspace{-0.5cm}
\end{figure}

We conduct several visualization studies on the COCO val2017 dataset~\cite{lin2014microsoft}. First, Fig.~\ref{fig:coco_distrib} shows the relative distribution of the embeddings produced by HyCoCLIP and our method, visualized using HoroPCA~\cite{chami2021horopca} according to their distance from the origin. HyCoCLIP embeddings lie closer to the origin, whereas ours are positioned farther from the origin in the hyperbolic space. In addition, our embeddings are more widely dispersed, with reduced overlap between part and whole image/text representations. This indicates that our hyperbolic alignment utilizes the available hyperbolic volume more effectively.

\begin{figure*}[!t]
\vspace{-0.5cm}
    \centering
     \includegraphics[width=0.93\textwidth]{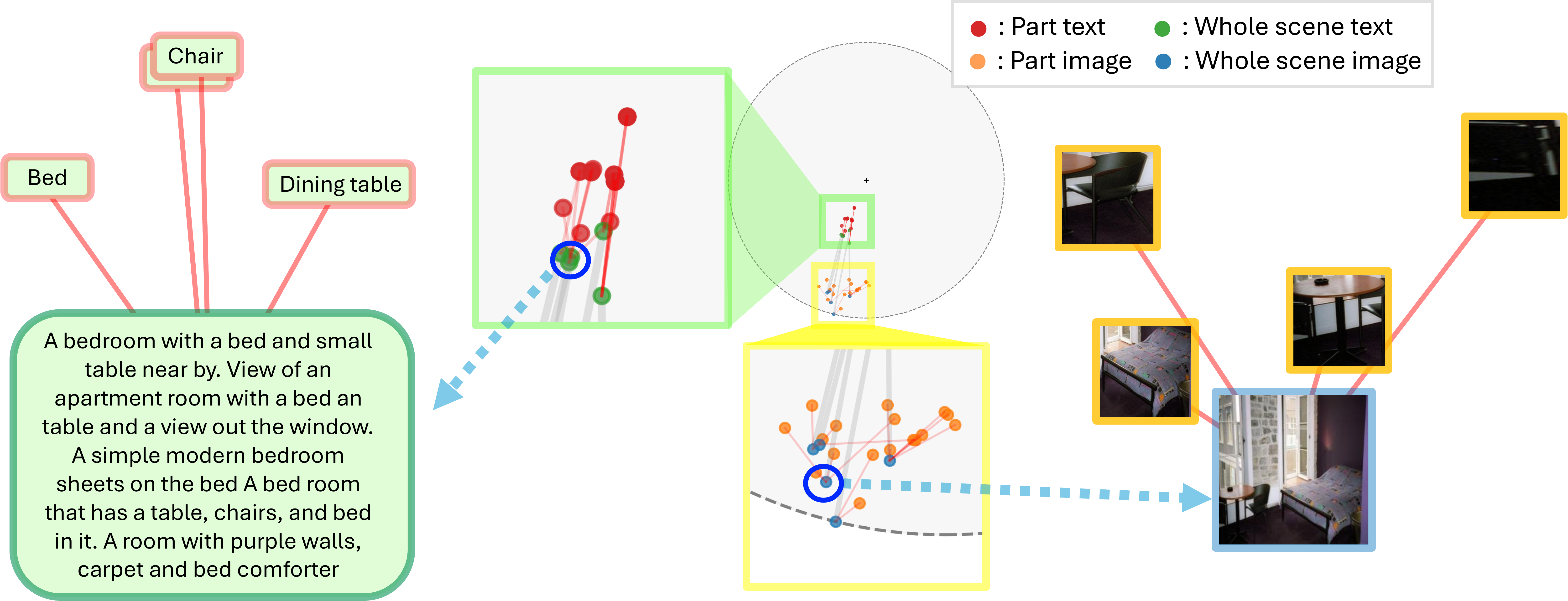}
    \caption{\textbf{Hyperbolic embedding of whole vs. part representations.}
Whole-scene images and texts lie deeper in the hyperbolic space, while part-level representations cluster closer to the origin. The zoom-in view and examples illustrate how parts such as chair, bed, and dining table are organized relative to the whole-scene embedding.}
    \label{fig:coco_vis}
    \setlength{\intextsep}{5pt} 
\end{figure*}

In addition, Fig.~\ref{fig:coco_vis} presents qualitative examples in which we visualize a subset of COCO part texts and part images using HoroPCA. As shown, the global image concept “bedroom” and its corresponding text representation reside farther from the origin in the hyperbolic space, while multiple part-level objects distribute across different regions according to their uncertainty. Note that the part text “chair” appears multiple times in the part-text dataset, so we depict its labels as stacked in the visualization. A similar pattern also emerges in the PCA visualization shown in the green box region of Fig.~\ref{fig:coco_vis}, where several part-text embeddings overlap due to the dataset.
 

\subsubsection{Hyperparameter sensitivity analysis}
We conduct an analysis on $\lambda_1$ and $\lambda_2$. Following prior studies on Leaky-ReLU activations~\cite{xu2015empirical}, we use a small $\alpha$ to preserve sufficient non-linearity while preventing unstable optimization. Results for $\lambda_1$ and $\lambda_2$ are summarized in Tab.~\ref{tab:ablation_compact}, where all models are trained for 100k iterations. For consistency, we follow the same training protocol and architectural setup as in our main experiments, using the ViT-S configuration.
In Tab.~\ref{tab:ablation_compact}, we report both classification (Cls.) and retrieval (Ret.) performance, where each value corresponds to the average over all classification and retrieval tasks, respectively. As shown in the table, our method consistently maintains stable performance across different choices of $\lambda_1$ and $\lambda_2$, with only minor variations. Notably, our approach tends to achieve either stronger classification or retrieval performance depending on the hyperparameter setting, while avoiding significant degradation in either metric. This demonstrates that our method is robust to the choice of hyperparameters and does not require sensitive tuning to achieve competitive performance.

\begin{table}[H]
\centering
{\fontsize{7pt}{10pt}\selectfont
\setlength{\tabcolsep}{6pt}
\renewcommand{\arraystretch}{0.9}
\setlength{\aboverulesep}{0pt}
\setlength{\belowrulesep}{0pt}
\caption{\textbf{Hyperparameter sensitivity analysis at 100k iterations.}Hyperparameter sensitivity analysis of $\lambda_1$ and $\lambda_2$ at 100k iterations. Each entry reports classification (Cls.) and retrieval (Ret.) performance averaged across all tasks. Our method demonstrates stable performance across a wide range of values, with $\lambda_1 = 0.5$ and $\lambda_2 = 10.0$ selected as the default setting.}

\begin{tabular}{@{}l@{\hspace{2pt}}ccccc}
\toprule
\multirow{2}{*}{$\boldsymbol{\lambda_1}$}
& {0.3} 
& {0.4} 
& \cellcolor{lightgreen}\textbf{0.5} 
& {0.6} 
& {0.7} \\
\cmidrule(lr){2-6}

& 31.9 / 63.6  
& 31.5 / 64.2
& \cellcolor{lightgreen} 31.6 / 63.8
& 31.5 / 64.2
& 31.1 / 63.4 \\
\midrule
\multirow{2}{*}{$\boldsymbol{\lambda_2}$}
& 9.0 
& 9.5 
& \cellcolor{lightgreen}\textbf{10.0} 
& 10.5 
& 11.0 \\
\cmidrule(lr){2-6}

& 31.3 / 64.2 
& 31.5 / 64.9 
& \cellcolor{lightgreen} 31.6 / 63.8
& 31.5 / 62.9
& 31.4 / 63.2 \\
\bottomrule
\label{tab:ablation_compact}
\end{tabular}
}
\vspace{-1.2em}
\end{table}

\subsubsection{Role and influence of individual loss terms}
\begin{figure*}
    \centering
    \includegraphics[width=0.93\textwidth]{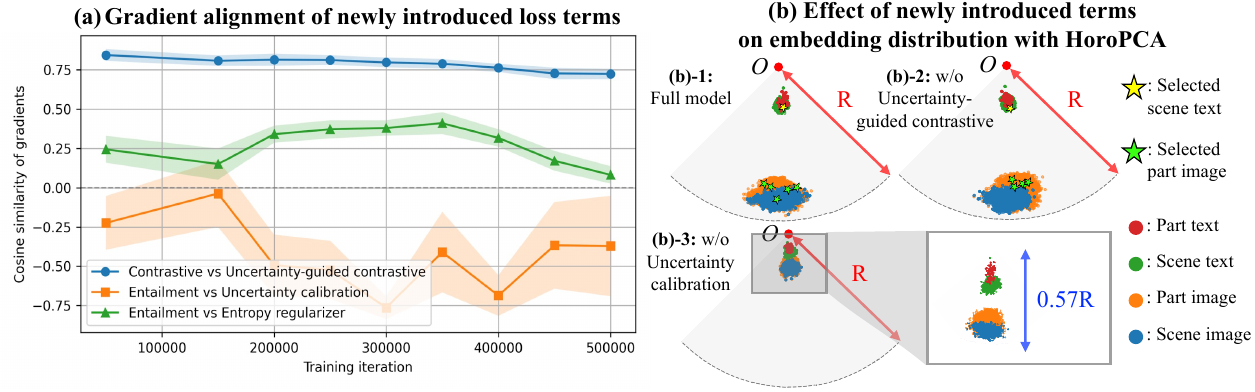}
    \caption{\textbf{Analysis of our newly introduced loss terms.} (a) Cosine similarity between gradients of different loss components, showing that the uncertainty calibration loss acts as a regularizer by opposing the entailment loss, while the uncertainty-guided contrastive loss remains aligned with the main contrastive objective. (b) Visualization of embedding distributions using HoroPCA on COCO, where the full model exhibits well-structured representations, while removing each loss term leads to degraded alignment or concentrated embeddings.}
    \label{fig:loss_influence}
\end{figure*}

We analyze the role of each loss component at 100k iterations. Fig.~\ref{fig:loss_influence}(a) shows the cosine similarity between gradients of different loss terms. The uncertainty calibration loss exhibits an opposing gradient direction to the entailment loss, acting as a regularizer that prevents representation collapse and stabilizes training. In contrast, the uncertainty-guided contrastive loss remains well aligned with the standard contrastive objective, reinforcing the primary learning signal.
Fig.~\ref{fig:loss_influence}(b) visualizes the embedding distributions on COCO~\cite{lin2014microsoft} using HoroPCA~\cite{chami2021horopca}. In the full model ((b)-1), embeddings are well-structured with clear relationships between scene text ($\textcolor{yellow}{\bigstar}$) and part images ($\textcolor{green}{\bigstar}$). Removing the uncertainty-guided contrastive loss ((b)-2) weakens this relational alignment, while removing the uncertainty calibration loss ((b)-3) causes the embeddings to concentrate in a narrower region (approximately $0.57R$), reducing representational capacity. Overall, the uncertainty-guided contrastive loss improves relational alignment, whereas the uncertainty calibration loss maintains a well-distributed embedding space and prevents such contraction.

\subsubsection{Embedding analysis on hyperbolic radius.}
\begin{figure*}
    \centering
    \includegraphics[width=0.6\textwidth]{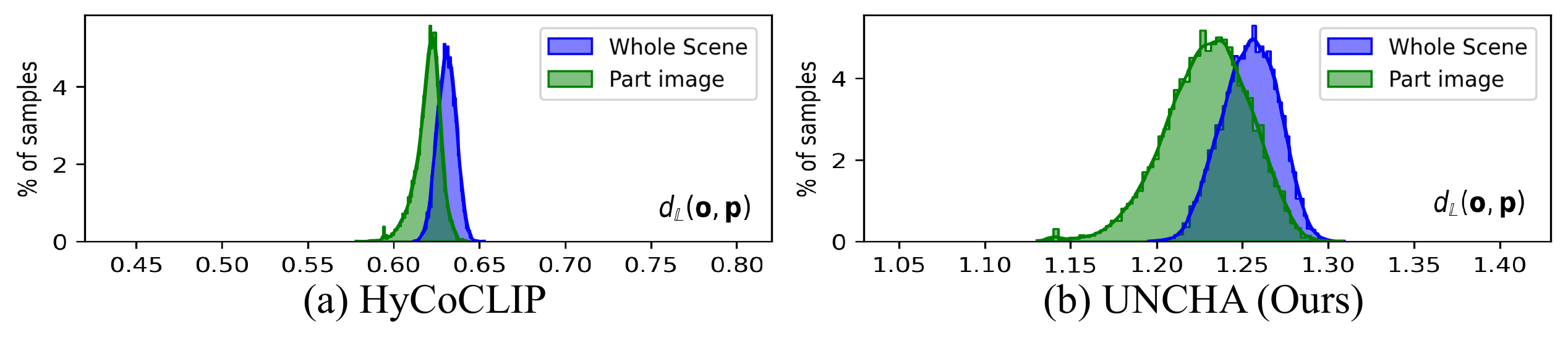}
    \vspace{-0.3cm}
    \caption{\textbf{Hyperbolic embedding analysis using hyperbolic radius.} Distances are measured by $d_{\mathbb{L}}(\mathbf{o}, \mathbf{p})$ instead of the Euclidean norm to better reflect the intrinsic geometry of hyperbolic space. The results show that embeddings are distributed across different radial regions, corresponding to varying levels of semantic granularity and representational capacity.}
    \label{fig:hyp_radius}
\end{figure*}

In Fig.4, following prior work~\cite{pal2024compositional}, we first visualize embedding distances using the Euclidean norm. However, this does not fully reflect the geometry of hyperbolic space. To address this, we re-plot the results using the hyperbolic distance from the origin, $d_{\mathbb{L}}(\mathbf{o}, \mathbf{p})$, in Fig.~\ref{fig:hyp_radius}.
Due to the exponential expansion of hyperbolic space with radius~\cite{bridson2013metric}, points farther from the origin lie in regions with significantly larger effective volume. Therefore, analyzing embeddings with $d_{\mathbb{L}}(\mathbf{o}, \mathbf{p})$ provides a more faithful view of their distribution and better captures hierarchical and semantic structure.

\subsubsection{Hyperbolic distribution during training}
\begin{figure*}[!t]
\vspace{-0.3cm}
    \centering
     \includegraphics[width=0.93\textwidth]{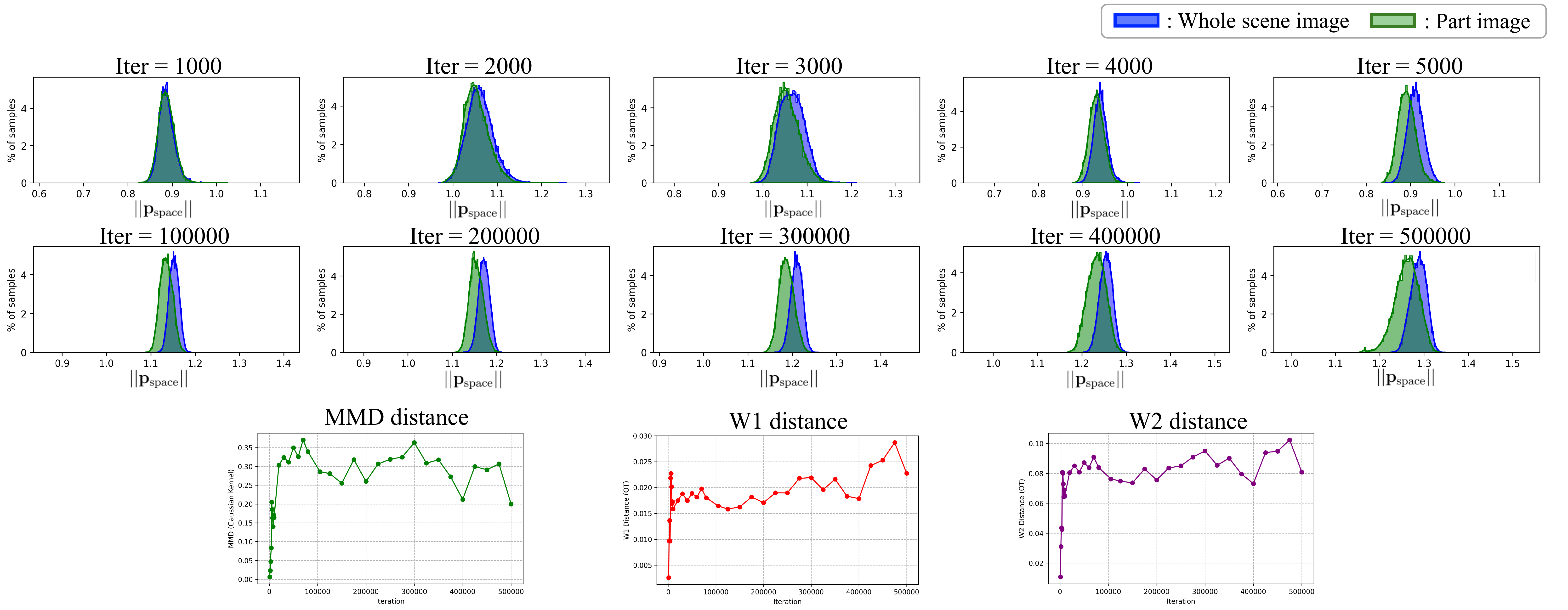}
     \vspace{-0.3cm}
    \caption{\textbf{Hyperbolic embedding distributions of whole images vs. part images across training iterations.} 
As training progresses, the uncertainty distributions of whole images and small crops gradually diverge, indicating increasing part–whole separation in the learned hyperbolic space. 
The bottom row reports iteration-wise distributional distances (MMD, W1, W2), which quantitatively confirm the growing discrepancy between the two distributions.}
    \label{fig:distrib_iter}
    \setlength{\intextsep}{5pt} 
    \vspace{-0.5cm}
\end{figure*}

To investigate how our hyperbolic alignment organizes part-whole relationships within the hyperbolic space, we visualize the distribution of embedding distances from origin for whole images and their corresponding part-level crops, using both cropped and full images from the ImageNet~\cite{deng2009imagenet, russakovsky2015imagenet} dataset. As shown in Fig.~\ref{fig:distrib_iter}, as training progresses, part-image distance from the origin decreases (\textit{i.e.}, the uncertainty associated with part images steadily increases), and the separation between the two distributions becomes more pronounced. This pattern indicates that the model gradually enhances its ability to distinguish part-level content from full-scene contexts.

The bottom row of Fig.~\ref{fig:distrib_iter} reports three statistical distances, Maximum Mean Discrepancy (MMD), Wasserstein-1 distance (W1), and Wasserstein-2 distance (W2), computed at every iteration, quantitatively confirming the growing divergence between the part and whole image distributions. Consistent with the visual trends, all three metrics rise sharply during the early stages of training and gradually stabilize as the model converges. W1 measures the minimum amount of mass that must be transported to align one distribution with the other, reflecting differences in their overall locations. W2 extends this by incorporating squared deviations, making it more sensitive to changes in distributional spread. MMD evaluates the discrepancy between two distributions by comparing their kernel-based mean embeddings, capturing differences in both central tendency and higher-order statistical structure.

\subsubsection{Dense feature localization visualization}
\begin{figure*}[!t]
    \centering
     \includegraphics[width=0.77\textwidth]{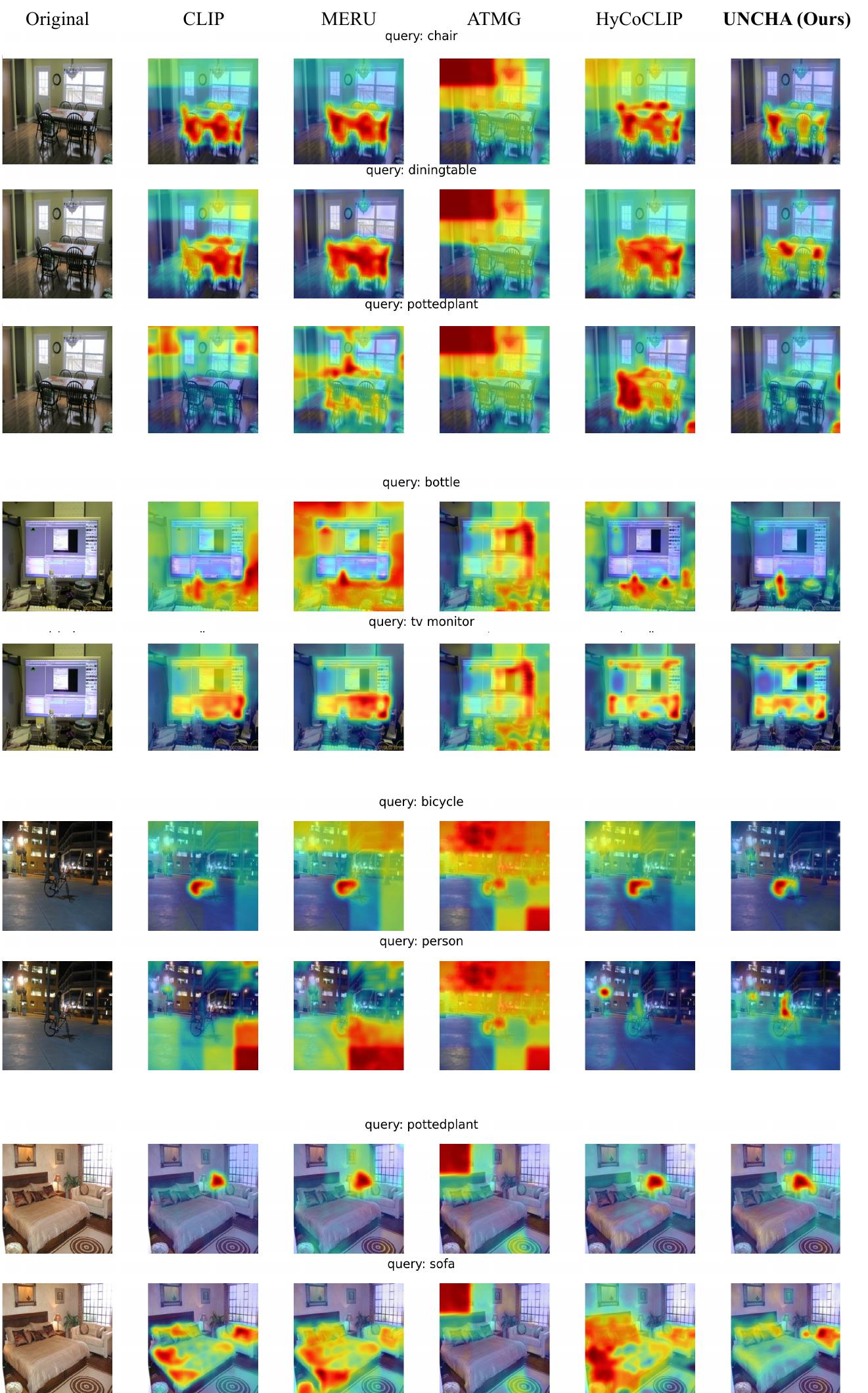}
     \vspace{-0.8em}
     \caption{\textbf{Dense feature localization visualizations for zero-shot semantic segmentation.}
Following the procedure described in Sec.~\ref{sec:semantic-seg}, similarity maps on the VOC dataset are generated by extracting dense features and computing their correspondence to text queries. Our method produces sharper and more localized activations that align more accurately with the queried object categories.}
    \label{fig:voc_vis_1}
       \vspace{-0.8em}
    \setlength{\intextsep}{5pt} 
\end{figure*}

\begin{figure*}[!t]
    \centering
     \includegraphics[width=0.77 \textwidth]{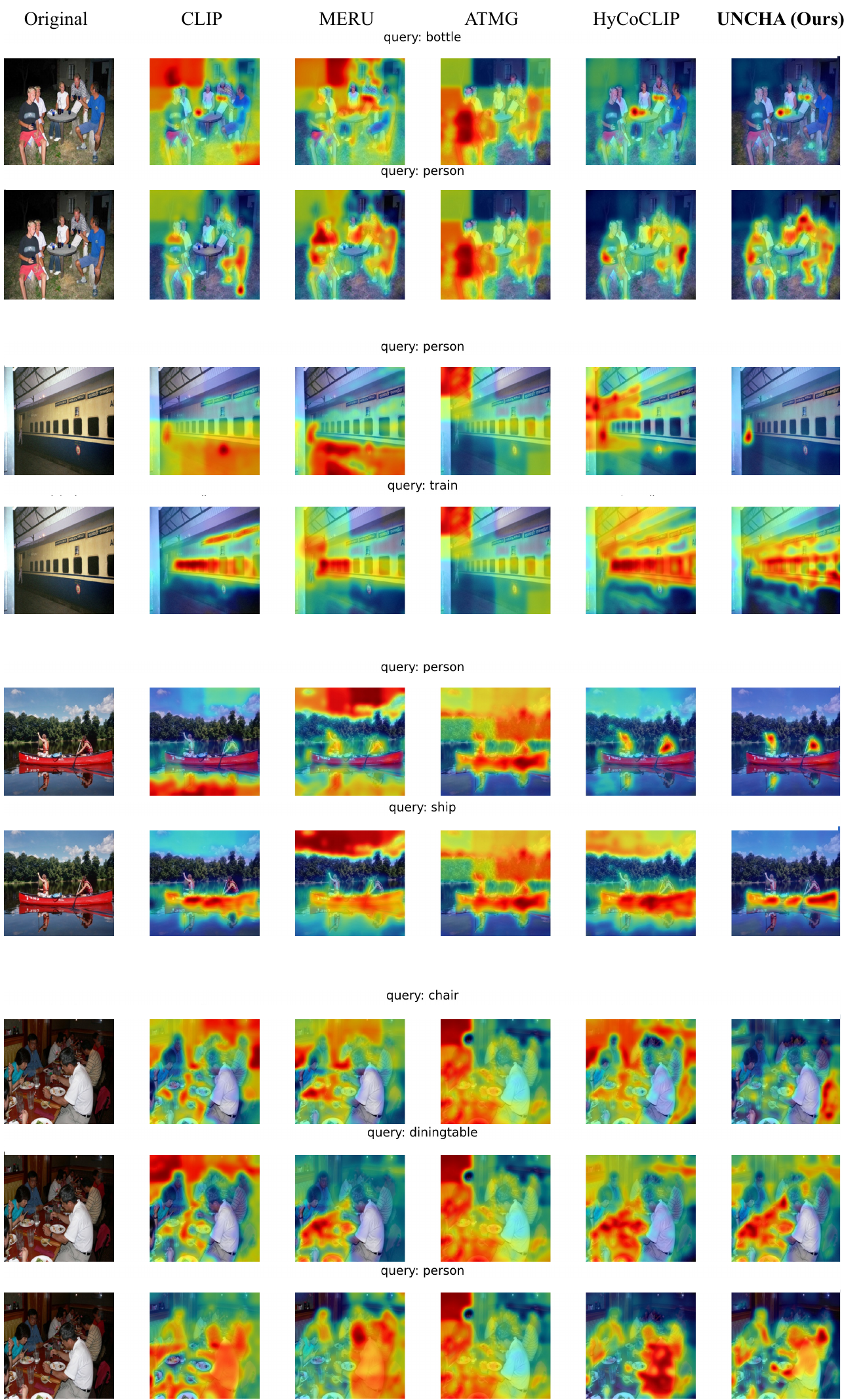}
     \vspace{-0.8em}
    \caption{\textbf{Dense feature visualizations for zero-shot semantic segmentation.}
Similarity maps on the VOC dataset are generated by extracting dense features and computing their correspondence to text queries, following the procedure described in Sec.~\ref{sec:semantic-seg}. 
Our method produces sharper and more localized activations that align more accurately with the queried object categories.}

    \label{fig:voc_vis_2}
       \vspace{-0.8em}
    \setlength{\intextsep}{5pt} 
\end{figure*}

    We follow a setting analogous to \ref{sec:semantic-seg} and perform dense localization on the VOC dataset~\cite{everingham2010pascal} by computing the similarity between text queries and dense features. The resulting visualizations are presented in Fig.~\ref{fig:voc_vis_1} and Fig.~\ref{fig:voc_vis_2}. As shown, our method consistently provides the most fine-grained and accurate localization across a diverse set of object classes and input images. Notably, our model is able to correctly highlight objects that competing methods either fail to capture (\textit{e.g.}, person, sofa) or detect with substantially less precision (\textit{e.g.}, dining table, potted plant). These findings demonstrate that our approach achieves a more detailed and robust understanding of complex, multi-object scenes compared to existing baselines. Quantitative results supporting these observations are reported in \ref{sec:semantic-seg}.


\subsubsection{Uncertainty-based ordering of part images}
\begin{figure*}[!t]
    \centering
     \includegraphics[width=0.6\textwidth]{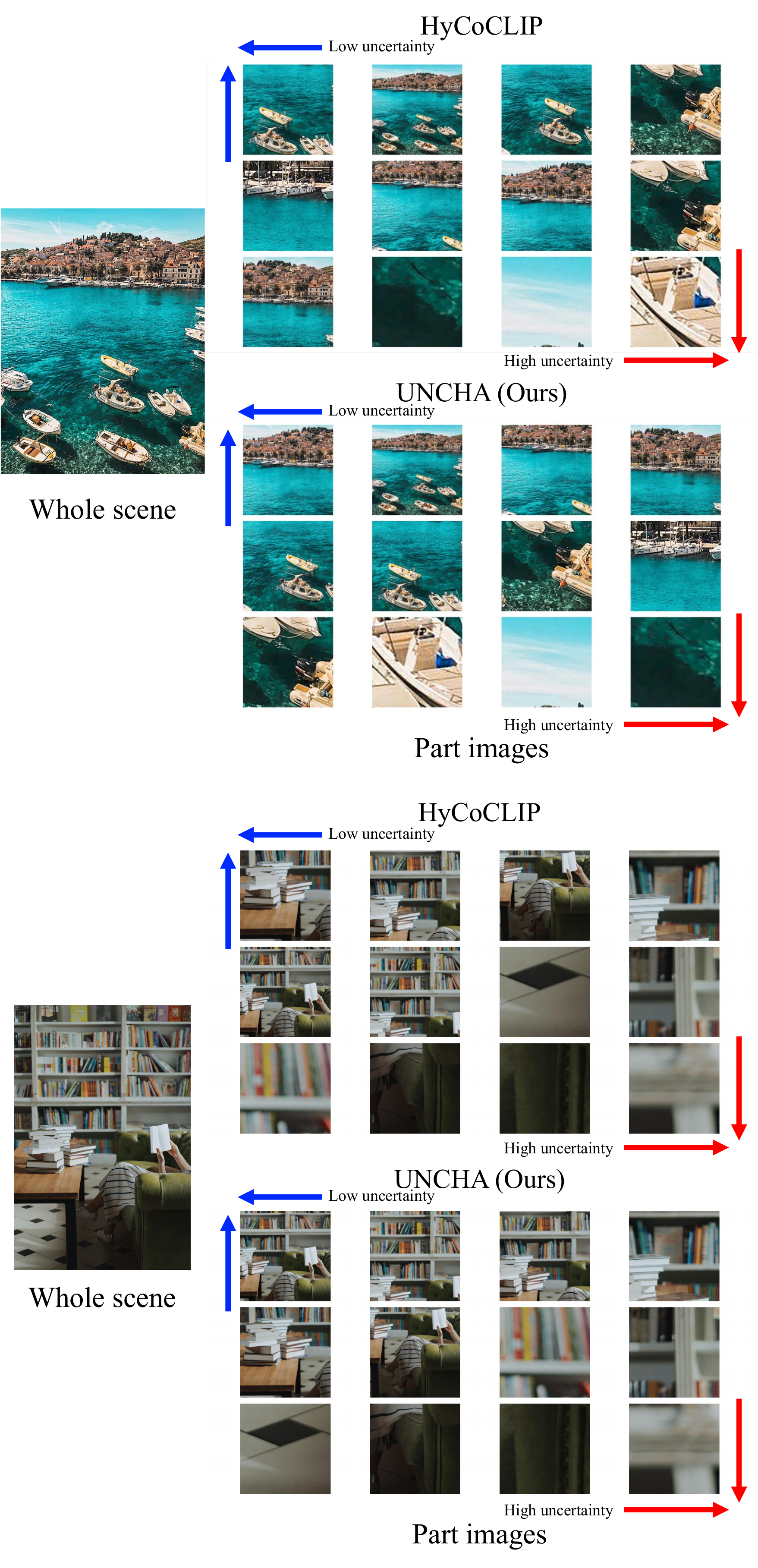}
     \vspace{-0.8em}
    \caption{\textbf{Comparison of uncertainty-based ordering of part images.} Comparison of uncertainty-based ordering of part images between HyCoCLIP~\cite{pal2024compositional} and UNCHA (Ours) shows that UNCHA produces a coherent ordering in which part images are arranged according to their scene-level representativeness.}
    \label{fig:norm-sorting-1}
       \vspace{-0.8em}
    \setlength{\intextsep}{5pt} 
\end{figure*}

We investigate how well part images are organized within the hyperbolic space by sorting them based on uncertainty and comparing them with HyCoCLIP. Because the Euclidean norm, hyperbolic radius, and uncertainty are monotonic measures (differing only in direction), we sort HyCoCLIP embeddings by their Euclidean norms for a fair comparison with our uncertainty-based ordering. The results are presented in Fig.~\ref{fig:norm-sorting-1}. As shown, HyCoCLIP produces several misordered cases where abstract or highly-representative samples appear in inconsistent positions, whereas our method yields a more coherent ordering in which part images align naturally according to their scene-level representativeness.

\subsubsection{Hyperbolic embedding visualization with various dataset}
\begin{figure*}[!t]
    \centering
     \includegraphics[width=0.93\textwidth]{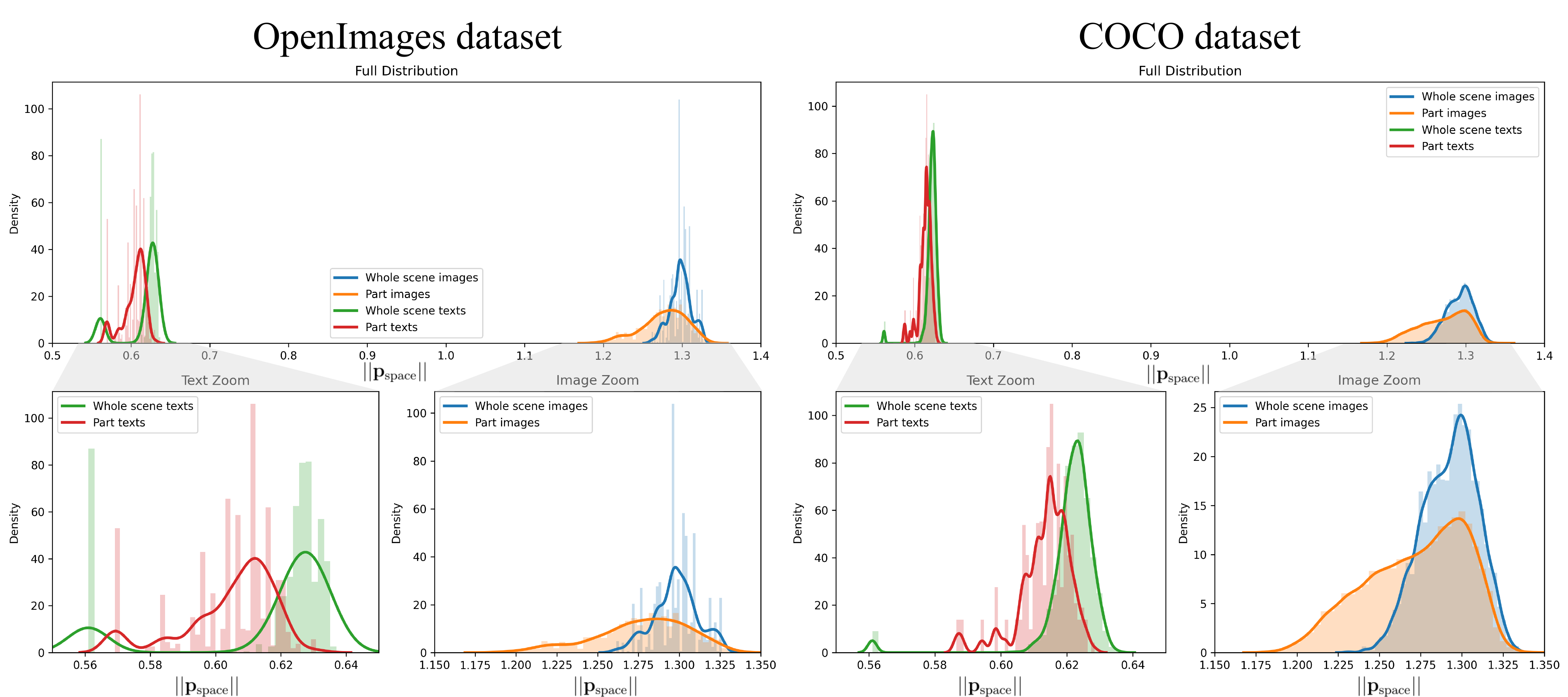}
    \caption{\textbf{Distribution of hyperbolic embeddings across datasets.}
Using UNCHA (ViT-B), we visualize part and whole representations from OpenImages~\cite{kuznetsova2020open} and COCO~\cite{lin2014microsoft} Across both datasets, part-level embeddings appear closer to the origin, while whole-scene embeddings lie farther away, consistently reflecting their hierarchical structure.}
    \setlength{\intextsep}{5pt} 
    \label{fig:norm_dist_dataset}
\end{figure*}

We analyze how part images, part texts, whole scene images, and whole scene texts are distributed within the hyperbolic embedding space by conducting the visualization shown in Fig.~\ref{fig:norm_dist_dataset}. All experiments are performed using our ViT-B model on both the COCO~\cite{lin2014microsoft} and OpenImages~\cite{kuznetsova2020open} datasets. As illustrated, part-level data consistently occupy regions closer to the origin compared to whole-scene representations, and this trend remains stable across different datasets.

\twocolumn
\clearpage
{
    \small
    \bibliographystyle{ieeenat_fullname}
    \bibliography{main,suppl}
}

\end{document}